%% file: arxiv.tex
\documentclass{article}

\PassOptionsToPackage{numbers, compress}{natbib}

\usepackage[margin=1in]{geometry}

\usepackage{microtype}      
\usepackage{xcolor}         

\usepackage[utf8]{inputenc} 
\usepackage[T1]{fontenc}    
\usepackage[colorlinks=true,linkcolor=blue,allcolors=blue]{hyperref}       
\usepackage{url}            
\usepackage{booktabs}       
\usepackage{amsfonts}       
\usepackage{nicefrac}       
\usepackage{microtype,nicefrac}      
\usepackage{xspace}
\usepackage{natbib}

\usepackage{comment}
\usepackage{amsmath}
\usepackage{amsthm}
\usepackage{graphicx}
\usepackage{rotating}
\usepackage{siunitx}

\usepackage{amsfonts}
\usepackage{amssymb}
\usepackage[overload]{empheq}
\usepackage{autobreak}
\usepackage{mathtools}
\usepackage{algorithm} 
\usepackage{algcompatible}
\usepackage[parfill]{parskip}
\usepackage[toc]{appendix}
\usepackage{minitoc}
\usepackage{enumitem}
\usepackage{multicol}
\usepackage{lipsum}
\usepackage{dsfont}
\usepackage{esint}

\input{macros-arxiv}

\definecolor{burntorange}{rgb}{0.8, 0.33, 0.0}
\newcommand{\sam}[1]{{ #1}}

\title{
Scale-Consistent Learning for Partial Differential Equations
}

\author{%
Zongyi Li, Samuel Lanthaler, Catherine Deng, Michael Chen,  Yixuan Wang,\\ 
Kamyar Azizzadenesheli$^\dagger$, Anima Anandkumar\\
\\
Caltech, $^\dagger$Nvidia
}

\date{}

\begin{document}

\maketitle

\begin{abstract}

\input{Sections/0-abstract}
\end{abstract}

\input{Sections/1-Intro-2025}
\input{Sections/2-RelatedWorks}
\input{Sections/3-Preliminaries}

\input{Sections/4-Algorithm}
\input{Sections/4-Method}

\input{Sections/5-Experiments}

\input{Sections/7-Conclusion}
\input{Sections/Acknowledgements}

\newpage
\bibliography{Main}
\bibliographystyle{unsrt}

\newpage
\appendix
\input{Sections/A-Data}
\input{Sections/A-Theory}
\input{Sections/A-Implementation}
\input{Sections/A-Experiments}

\end{document}

%% file: macros-arxiv.tex
\newcommand{\G}{\mathcal{G}}
\newcommand{\cG}{\G}
\newcommand{\cP}{\mathcal{P}}
\newcommand{\cB}{\mathcal{B}}

\newcommand{\U}{\mathcal{U}}

\newcommand{\T}{\mathcal T}

\newcommand{\R}{\mathbb{R}}

\newtheorem{lemma}{Lemma}[section]
\newtheorem{remark}{Remark}[section]
\newtheorem{proof*}{Proof}[section]

\newtheorem{definition}{Definition}[section]
\newtheorem{theorem}{Theorem}[section]

%% file: Sections/0-Abstract.tex
Machine learning (ML) models have emerged as a promising approach for solving partial differential equations (PDEs) in science and engineering. Previous ML models typically cannot generalize outside the training data; for example, a trained ML model for the Navier-Stokes equations only works for a fixed Reynolds number ($Re$) on a pre-defined domain. 
To overcome these limitations,  we propose a data augmentation scheme based on scale-consistency properties of PDEs and design a scale-informed neural operator that can model a wide range of scales. 
Our formulation leverages the facts: (i) PDEs can be rescaled, or more concretely, a given domain can be re-scaled to unit size, and the parameters and the boundary conditions of the PDE can be appropriately adjusted to represent the original solution, and (ii) the solution operators on a given domain are consistent on the sub-domains. We leverage these facts to create a scale-consistency loss that encourages matching the solutions evaluated on a given domain and the solution obtained on its sub-domain from the rescaled PDE. 
Since neural operators can fit to multiple scales and resolutions, they are the natural choice for incorporating scale-consistency loss during training of neural PDE solvers. 
We experiment with scale-consistency loss and the scale-informed neural operator model on the Burgers' equation, Darcy Flow, Helmholtz equation, and Navier-Stokes equations.
With scale-consistency, the model trained on $Re$ of 1000 can generalize to $Re$ ranging from 250 to 10000, and reduces the error by 34\% on average of all datasets compared to baselines.


%% file: Sections/1-Intro-2025.tex
\begin{figure}[t]
\begin{center}
\includegraphics[width=0.8\columnwidth]{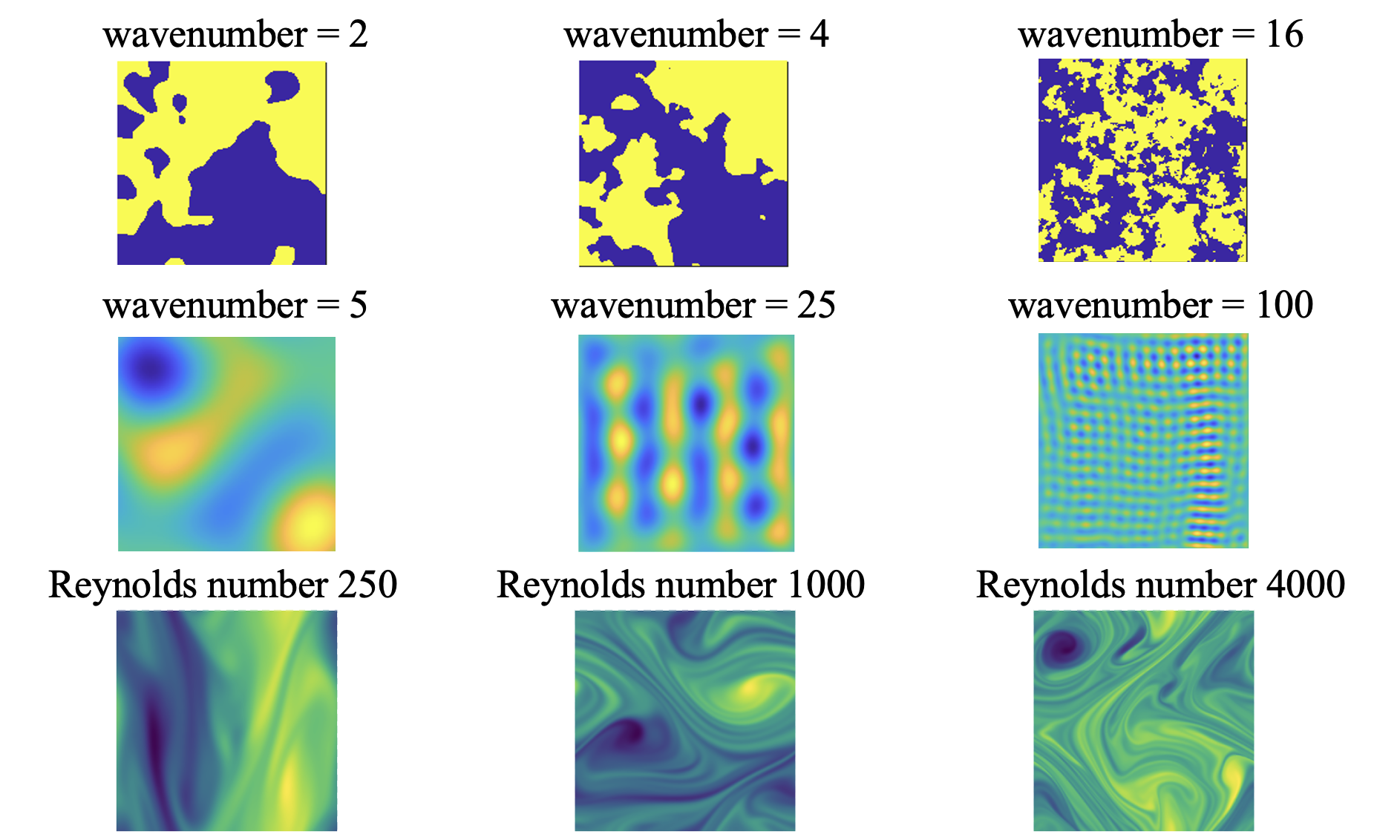}
\caption{ Multi-scale PDE dataset:
Continuum mechanics at different scales (kilometer- or millimeter-scale) can be formulated to a unit-scaled domain with corresponding scale parameters.
\textbf{Row 1}: Darcy Flows,
\textbf{Row 2}: Helmholtz Equation, 
\textbf{Row 3}: Navier Stokes equation.
In this work, we aim to design a learning framework to capture the consistency across the scales.
}
\label{fig:dataset}
\end{center}
\end{figure}

\begin{figure*}[t]
\centering
\includegraphics[width=\textwidth]{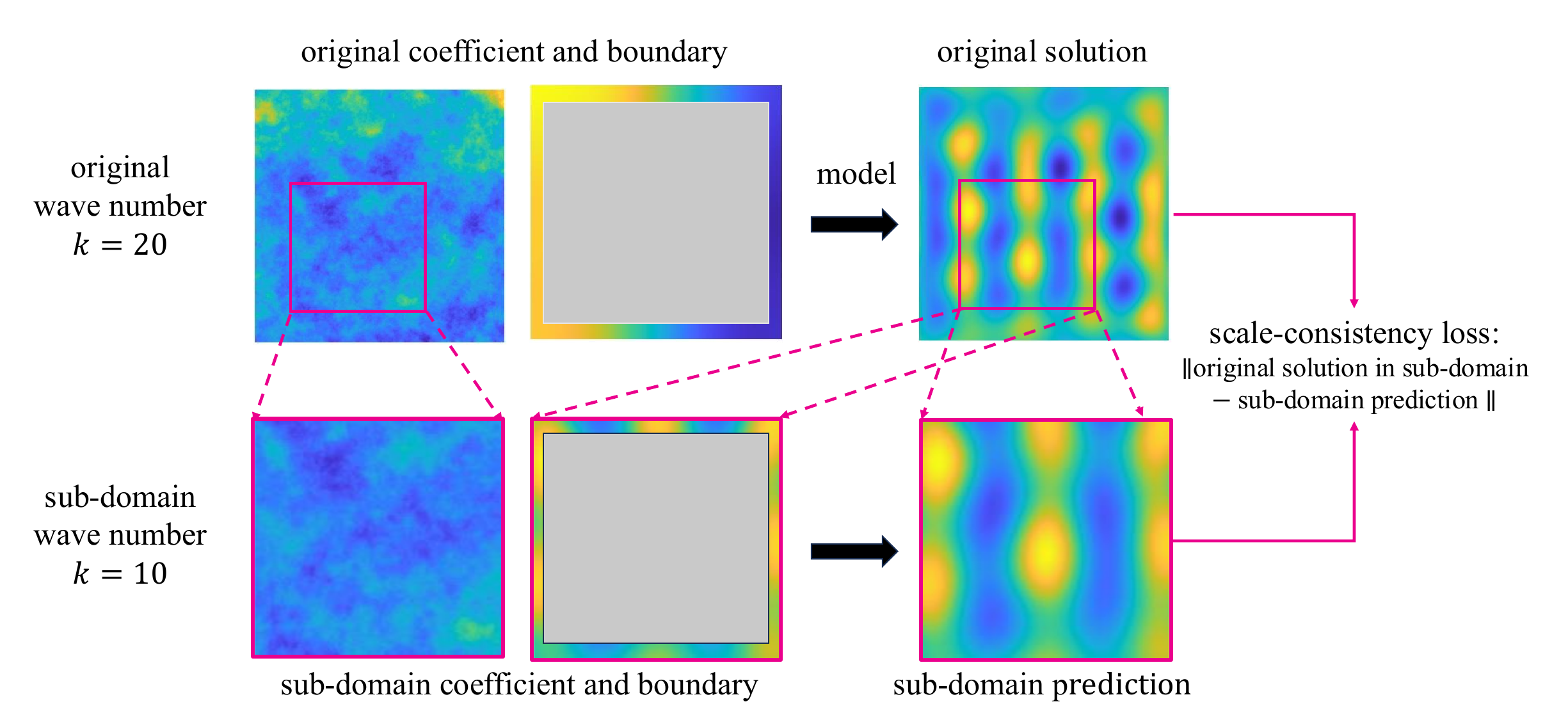}

\caption{Scale-consistency loss is achieved via sub-domain sampling and re-scaling. Given a data instance consisting of an input coefficient, boundary, scale parameter, and solution, we restrict these elements to a sub-domain. This process creates a new data instance that has the same resolution but a smaller grid size. The sub-domain is then rescaled to a unit length while the grid sizes are kept unchanged. The scale-consistency loss is defined as the discrepancy between the global and sub-domain predictions.
}
\label{fig:algo}
\end{figure*}

\section{Introduction}

\textbf{ML for PDEs.}
Data-driven methods have become increasingly popular in learning Partial Differential Equations (PDEs) for scientific computing \cite{azizzadenesheli2024neural}, showing various applications ranging from weather forecasting \cite{pathak2022fourcastnet} to nuclear fusion \cite{gopakumar2023fourier}. 
While conventional models are typically parameterized for a fixed resolution at a predefined scale, neural operators have recently been proposed to generalize across discretization by parameterizing the model on function spaces \cite{li2020neural,kovachki2023neural,Lu_2021,kissas2022learning,raonic2023convolutional}.
Among these, the Fourier Neural Operator (FNO) \cite{li2020fourier} stands out as one of the most efficient models. It learns dynamics on the frequency domain, which can be viewed as an efficient, resolution-invariant tokenization. Recent advances further improve the model with shared kernel \cite{AFNO} and  U-shape architectures  \cite{uno}.
Given promising results, one of the major challenges of scientific machine learning has been the lack of high-quality training data.





\textbf{Self-supervised learning for PDEs.}
To overcome the limitation of data, many self-supervised learning techniques have been studied. Especially, the AI for Science community has investigated building-in physics knowledge to the models via equation loss \cite{li2021physics} and symmetry augmentation \cite{wang2020incorporating, brandstetter2022lie}. For two-dimensional PDEs, the symmetry groups include translation, rotation, Galilean boost, and scaling. Among them, scale symmetry has been the least effective in improving performance~\cite{brandstetter2022lie,mialon2023self}. Our hypothesis is that previous formulations of scale-symmetry are defined as positional encoding, which does not incorporate scaling parameters and boundary conditions. 


\textbf{Multi-scale behavior in physics}. Many natural phenomena exhibit multiscale behavior, i.e., interact across a wide range of scales.  
This is especially the case with solutions of partial differential equations (PDEs), which model various phenomena in science and engineering. For instance, the Navier-Stokes equation, a classical model describing fluid motion, applies to kilometer-scale problems such as weather forecasting \cite{pathak2022fourcastnet}, meter-scale problems such as airfoils \cite{ginopaper}, and millimeter-scale problems such as catheters \cite{zhou2023ai}. 

\textbf{PDEs can be rescaled.}
While the physics at the kilometer and millimeter scales exhibit very different behaviors and frequency ranges, continuum mechanics can be universally reformulated in PDEs using scale parameters, such as the Reynolds number in the Navier-Stokes equation, as illustrated in figure~\ref{fig:dataset}. 
\begin{definition}[Rescaling of PDEs]
In general, a PDE $R$ with coefficient function $a$ and solution function $u$ on domain $\Omega$
\begin{equation}
    \label{eq:intro-darcy}
    \begin{split}
            R\left( a(x), u(x)\right) = 0, \quad\quad (x\in \Omega)
    \end{split}
\end{equation}
can be rescaled to a new domain size $\Omega_\lambda$ with scaling $\lambda$,
\[R_{\lambda} \left( a(\lambda x), u(\lambda x)\right) = 0, \quad\quad (x\in \Omega_\lambda)\]
\end{definition}
For example, in the Darcy flow, $R(a, u) = \nabla (a \nabla u)$. Rescaling to the unit domain is also called nondimensionalization.

Further, the solution operators on a given domain are consistent on the sub-domains. Thus, given a domain, the values of the PDE solution in a subdomain can be equivalently obtained by rescaling the subdomain to unit size but by choosing a different set of appropriate parameters (known as scale parameters) and boundary conditions in the PDE. 


\textbf{Our approach.} 
Based on the above observation,  we define the scale-consistency loss as the overall difference between the original solution of the PDE limited to the subdomain, and the one obtained from the modified PDE upon rescaling the subdomain to unit size, as shown in Figure \ref{fig:algo}. The ground-truth solution has a zero scale-consistency loss, or in other words, solution operators of PDEs are scale consistent. 


We apply scale-consistency loss as a data augmentation procedure during the training of neural operators for solving PDEs. We address the challenging task of modeling PDEs that exhibit dramatically different behaviors across scales. Our new dataset consists of four PDEs at different scales: the Darcy flow with varying coefficients, the Burgers' equation with viscosity ranging from $1/100$ to $1/1000$, the Helmholtz equation with wavenumbers spanning $1$ to $100$, and the Navier-Stokes equation with Reynolds numbers from $250$ to $10000$, as shown in Figure \ref{fig:dataset}. To evaluate generalization capabilities, we train models at specific scales and test their performance across different scales. In particularly challenging cases, such as the Helmholtz equation, different wavenumbers result in entirely distinct frequency ranges, causing all baseline models to fail at generalization. However, by incorporating our scale-consistency loss, the model successfully achieves zero-shot extrapolation to previously unseen scales, i.e., PDEs with scale parameters not available during training. Our main contributions are as follows.
\begin{itemize}[leftmargin=0.3cm]
  \setlength\itemsep{-.2em}
    \item We propose a data augmentation scheme based on scale-consistency loss that creates data instances with various scales via sub- and super-sampling. For time-dependent problems, we sample in the space-time domain.
    \item We show a theorem (\ref{thm:sc}) for elliptic PDEs that, under mild assumptions, low scale-consistency loss guarantees recovery of the underlying solution operator.
    \item We design a scale-informed neural operator that takes the scale parameter as input with weight-sharing parameterization and adaptive U-shape architecture to capture a wide range of scales.
    \item Based on the pre-trained scale-consistent neural operator, we propose an domain-decomposition algorithm to iteratively refine the output at test time, which further reduces the error by 40\% on the Darcy Flow.
    \item We propose a challenging multiscale dataset including the Burgers' equation, Darcy Flow, Helmholtz equation, and Navier-Stokes equation. The results show that the scale-consistency loss helps the scale-informed neural operator extrapolate to wider scales with a $34\%$ error reduction on average compared to baseline FNO models at the cost of double runtime.
\end{itemize}



To capture a wide range of scales, we propose a new architecture named the scale-informed neural operator, as shown in Figure \ref{fig:architecture}. 
We use the Fourier neural operator (FNO)~\cite{li2020fourier} as the backbone, as FNO naturally handles varying resolution by mapping inputs to the Fourier basis of unit domain size.
We incorporate the scale parameter as an additional input and embed the scale features in the frequency space, helping the model capture different frequencies corresponding to different scale parameters.
Inspired by \cite{AFNO}, we use a weight-sharing parameterization, where a single weight network is shared across all frequency modes. Additionally, it employs a multi-band U-shaped architecture similar to \cite{uno} that optimizes channel dimensions, using larger dimensions for lower frequency bands and smaller dimensions for higher frequency bands. 

Once trained, our scale-consistent neural operator can be deployed at inference time using a domain decomposition scheme. As detailed in Section \ref{algo:domain-decomp}, the process begins by predicting a coarse, global solution. This solution is then used to initialize local refinements on subdomains, which are processed by the same pre-trained operator. This approach provides two key advantages over standard domain decomposition. First, by initializing the subdomain problems with an informed coarse prediction rather than with zeros, our method converges significantly faster. Second, our model facilitates a multi-level, coarse-to-fine refinement strategy, breaking the two-scale limitation of conventional methods that often struggle with high iteration counts when the target scale is very fine. This test-time domain decomposition algorithm reduces the final prediction error by an additional 40\%.


%% file: Sections/2-RelatedWorks.tex
\section{Related Work}

\paragraph{Neural operator and foundation models.}
Data-driven models have become a common methodology to complement or augment numerical solvers for physical simulation \cite{wang2023scientific}. However, existing data-driven models are typically targeted to a single input variable, such as the coefficient function or initial condition, while other parameters remain fixed, including the domain size, boundary condition, and forcing term \cite{takamoto2022pdebench}. Recently, foundation models have been proposed to capture various datasets under a wide range of conditions, or even multiple families of PDEs \citep{subramanian2024towards,mccabe2023multiple, hao2024dpot,shen2024ups,rahman2024pretraining}. 
However, they do not explicitly capture relationships across a wide range of scales seen in physical systems. 
It is challenging for standard neural networks to capture different scales. 
In general, separate neural network models are trained for capturing each scale, making it cumbersome to couple them together and impose constraints across scales.



\paragraph{Symmetry-based augmentation.}
Scaling symmetry has been explored as a data augmentation technique in several works \cite{wang2020incorporating, brandstetter2022lie, mialon2023self}. In dynamical systems, this symmetry represents a fundamental relationship between spatial coordinates, time evolution, and field magnitudes. However, both \cite{brandstetter2022lie} and \cite{mialon2023self} reported limited effectiveness of this approach. This limitation may stem from two key challenges: first, continuous scaling symmetry becomes ill-defined on periodic domains without boundaries \cite{brandstetter2022lie}, and second, scaling velocity magnitudes disrupts the natural range of the input space. This is particularly problematic in applications like weather forecasting, where velocity fields typically maintain consistent magnitude ranges.
To address these limitations, we propose a generalized scaling consistenct framework that explicitly incorporates scaling parameters and boundary condition. 

\paragraph{Nondimensionalization and homogenization.}
The concept of rescaling has been widely applied in numerical partial differential equations. PDEs arising in physics and engineering are usually rescaled to a domain of unit size, omitting physical units in a process called nondimensionalization. A scale parameter that arises from this process, such as the Reynolds number in the Navier-Stokes equations, is called a dimensionless parameter. In the proposed scale-informed neural operator, a dimensionless parameter like the Reynolds number is provided as an input to the model to inform the scale. Separately, the coarse-graining of oscillating coefficients, such as conductivity or permeability, is called homogenization \cite{pavliotis2008multiscale}, where the local coefficient is averaged, leading to a simplified system. Previous work has shown success in learning constitutive laws for elliptic operators \cite{bhattacharya2024learning}, but hyperbolic equations in fluid mechanics remain challenging, as the scales cannot be easily separated.

\paragraph{Domain decomposition methods for partial differential equations.}
Domain decomposition (DD) is a class of methods for solving partial differential equations (PDEs) with multiscale coefficient functions \cite{smith1997domain, chen2021exponential}. This approach decomposes the domain into smaller subdomains, leading to simpler and more uniform local coefficient functions. DD is especially effective for elliptic boundary value problems, as the solution operator is linear with respect to the boundary function. Similar ideas have been applied to physics-informed neural networks \cite{jagtap2020extended} and other machine learning-based PDE solvers \cite{wang2022mosaic, huang2025operator}. However, previous works on neural operators usually require pre-determined domain sizes and scale parameters. In this work, we propose a learning algorithm to train a universal neural operator for various scales, which can be applied with domain decomposition at test time. With the pre-trained scale-consistent operators, we can initialize the interior with coarse prediction, and decompose the domains with multi-level domain sizes.





%% file: Sections/3-Preliminaries.tex
\section{Scale Consistency}
Many PDEs possess symmetries, which are reflected by the fact that the equations remain invariant under transformations such as translation, rotation, or re-scaling. An example is the Darcy flow problem on a domain $\Omega \in \R^d$.
\begin{subequations}
\label{eq:darcy}
\begin{align}[left = \empheqlbrace\,]
-\nabla \cdot \left( a(x) \nabla u(x)\right) &= 0, \quad\quad (x\in \Omega), \label{eq:d1} \\
u(x) &= g(x), \quad (x\in \partial \Omega). \label{eq:d2}
\end{align}
\end{subequations}
The associated solution operator $\cG$ is defined as a mapping
\[
(a(x),g(x)) \mapsto \cG(a,g)(x) := u(x).
\]


\subsection{Scale symmetry and scale consistency}

\textbf{Re-scale symmetry.}
Let $\T_\lambda$ be the re-scaling operator with $\lambda \in \R^+$ defined by $(\T_\lambda a)(x) := a(\lambda x)$ (or more generally with translation $ (\T_\lambda a)(x) = a(\lambda x + b)$ with $b \in \R^d$). In the absence of boundary conditions, the scale symmetry implies an equivariance property of $\cG$: 
\[
\cG(\T_\lambda a, \ldots) = \T_\lambda\cG(a,\ldots).
\]
The boundary condition (or simply the fact that the PDE is defined on a bounded domain $\Omega$) breaks the scale symmetry; if $u: \Omega \to \R$ is defined on the domain $\Omega$, then $\T_\lambda u$ is defined on the rescaled domain 
$
\Omega_\lambda = \{\lambda^{-1} x|x\in \Omega\},
$
 and we are generally lacking information about the boundary condition of the re-scaled domain $\partial \Omega_\lambda$. Thus, the presence of boundaries in most problems of practical interest makes it difficult to leverage the underlying symmetry properties of the equations in a straightforward way.


Nevertheless, under some conditions on the domain $\Omega$ (e.g. $\Omega = [0,1]^d$ is a cube), the formal scale symmetry of the solution operator of \eqref{eq:darcy} implies that if $u(x)$ solves \eqref{eq:darcy} with coefficient field $a(x)$ and with boundary condition $g(x)$, then the rescaled function $u_\lambda(x) = \T_\lambda u(x) = u(\lambda x)$, solves
\begin{align*}[left = \empheqlbrace\,]
-\nabla \cdot \left( a_\lambda(x) \nabla u_\lambda(x)\right) &= 0, \quad\quad (x\in \Omega_\lambda), \\
u_\lambda(x) &= \T_\lambda u(x), \quad (x\in \partial \Omega_\lambda). 
\end{align*}

i.e. $u_\lambda(x)$ is a solution of the Darcy flow problem on domain $\Omega_\lambda$, with coefficient field $a_\lambda = \T_\lambda a$, and boundary condition $(\T_\lambda u) |_{\partial \Omega_\lambda}$. Another operation we can perform is the restriction from $\Omega_\lambda$ to $\Omega$ when $\lambda\le1$. Intuitively, this condition expresses the fact that the solution operator of \eqref{eq:darcy} is  \textbf{scale-consistent}: The solution on a smaller subdomain $\Omega \subset \Omega_\lambda$ can either be obtained 
\begin{enumerate}[leftmargin=0.5cm]
\item by solving the PDE over the entire domain $\Omega_\lambda$ and then restricting the solution $u$ to the smaller domain $u|_{\Omega}$.
\item by solving the PDE directly on the subdomain $\Omega$, and imposing consistent boundary condition $u|_{\partial \Omega}$.
\end{enumerate}
Combining the scale symmetry with restriction, we obtain a new equation \eqref{eq:darcy} corresponding to the sub-domain of the original equation \eqref{eq:darcy-sub}. 
\begin{subequations}
\label{eq:darcy-sub}
\begin{align}[left = \empheqlbrace\,]
-\nabla \cdot \left( a_\lambda(x) \nabla u_\lambda(x)\right) &= 0, \quad\quad (x\in \Omega), \\
u_\lambda(x) &= \T_\lambda u(x), \quad (x\in \partial \Omega). 
\end{align}
\end{subequations}

By uniqueness of the equation, the solution of \eqref{eq:darcy-sub} must be consistent with the original solution in \eqref{eq:darcy}.
\begin{lemma}[Scale-consistency (solution function)]
\label{thm:sc}
If a function $u$ satisfies equation \eqref{eq:darcy}, then $u_\lambda = \T_\lambda u$ is the unique solution of equation \eqref{eq:darcy-sub}. 
\end{lemma}

Therefore, we obtain the following identity in terms of the solution operator $\cG$: let $\lambda \le 1$
\begin{align}
\begin{split}
\label{eq:scaling}
[\T_\lambda \cG(a,g)]|_{ \Omega}&
= \cG([\T_\lambda a]|_{ \Omega} , [\T_\lambda u]|_{\partial \Omega} ) \\
&\equiv \cG([\T_\lambda a]|_{ \Omega} , [\T_\lambda \cG(a,g)]|_{\partial \Omega} )
.
\end{split}
\end{align}
For the solution operator, this identity holds for arbitrary inputs $a(x)$ and $g(x)$. The scale-consistency \eqref{eq:scaling} can be used as a loss to train solution operators. Informally, if an operator satisfies \eqref{eq:scaling}, then it must be the target solution operator. The proof can be found at \ref{sec:proof}.

\begin{theorem}[Scale-consistency (solution operator)]
\label{thm:sc}
 If an operator $\Psi$ satisfies the scale-consistency \eqref{eq:scaling} and it matches the ground truth solution operator $\cG$ on nearly constant coefficient functions, then $\Psi \equiv \cG$.
\end{theorem}


\textbf{Scale-consistency loss.}
The first way to impose such a constraint is by introducing a loss of the form 
\begin{equation}
\label{eq:sc}
L(a, g) = 
\Vert \T_\lambda \Psi(a,g) - \Psi(\T_\lambda a, \T_\lambda \Psi(a,g)|_{\partial \Omega} ) \Vert.
\end{equation}

Note that this is an self-supervised loss term that doesn't require access to labeled data $u=\cG(a,g)$. It only requires producing input function samples $(a,g)$.
When solution data $u$ is available, the scale-consistency loss simplifies to 
\begin{equation}
\label{eq:aug}
L(a, g) = 
\Vert \T_\lambda u - \Psi(\T_\lambda a, \T_\lambda u|_{\partial \Omega} )   ) \Vert.
\end{equation}

\textbf{Infinitesimal scale-consistency.}
Another way to impose this constraint is by taking the $\lambda$-derivative of \eqref{eq:scaling}, leading to:
\[
\partial_\lambda \T_\lambda \cG(a,g) = \partial_\lambda\left[ \cG(\T_\lambda a, \T_\lambda \cG(a,g)|_{\partial \Omega})\right].
\]
We note that if $a(x)$ is a function, then the derivative $\partial_\lambda \T_\lambda a$ evaluated at $\lambda = 1$, is given by 
\[
\partial_{\lambda} \T_\lambda a|_{\lambda = 1} = \left[\partial_{\lambda} a(\lambda x)\right]_{\lambda=1} = x \cdot \nabla a(x),
\]
i.e., a radial spatial derivative of $a$. Substitution of this identity, and noting that $\T_{\lambda=1} a = a$ and $\T_{\lambda=1} \cG(a,g)|_{\partial \Omega} = g$, implies that 
\begin{align*}
\begin{split}
&x \cdot \nabla_x[\cG(a,g)](x) \\
&= \left\langle \frac{\delta \cG(a,g)}{\delta a}, x\cdot \nabla_x a \right\rangle + \left\langle \frac{\delta \cG(a,g)}{\delta g}, x\cdot \nabla_x [\cG(a,g)] \right\rangle
\end{split}
\end{align*}
We observe that while \eqref{eq:scaling} is highly non-linear, the infinitesimal constraint is quadratic in $\cG$.


\subsubsection{Scale-dependent problem: extension beyond scale symmetry}
The scale-consistency constraint can be written in greater generality, even if the underlying PDE has no scale symmetry. In this case, the domain could be an input to the operator, and the relevant scale-consistency would be 
\[
\cG(a,g;\Omega)|_{\Omega'} = \cG(a|_{\Omega'}, \cG(a,g,\Omega)|_{\partial \Omega'}; \Omega'),
\quad (\Omega' \subset \Omega).
\]
\sam{In some cases, this is equivalent to scaling certain parameters in the PDE, as explained below.}

\textbf{Helmholtz equation.}
An example not satisfying scale symmetry is the Helmholtz equation, 
\begin{equation}
\label{eq:helmholtz}
   -\nabla \cdot \left( a(x) \nabla u(x)\right) + k^2 u(x) = f(x).
\end{equation}
In this case, a rescaling of the spatial variable corresponds to a rescaling of the frequency $k^2$, i.e. $u_\lambda(x) = u(\lambda x)$ solves
$
-\nabla \cdot \left( a_\lambda(x) \nabla u_\lambda(x)\right) + \lambda^{-2} k^2 u_\lambda(x) = \lambda^{-2} f(\lambda x),
$
or 
\[
-\nabla \cdot \left( a_\lambda(x) \nabla u_\lambda(x)\right) + k^2_\lambda u_\lambda(x) = f_\lambda(x),
\]
with 
$
k_\lambda := \lambda^{-1} k, f_\lambda(x) := \lambda^{-2} f(\lambda x).
$
Thus, the scale-consistency constraint involves the whole family of PDEs,
$
\Delta u + k^2 u = f,
$
for $k>0$, with the transform on parameter $\T_\lambda (k) = \lambda k$.

\subsubsection{Time-dependent problem: rescale in space-time domain}
For time-dependent problems, in general, we could view the time dimension as another spatial dimension, and rescale both the spatial and temporal dimensions.

\textbf{Navier-Stokes equation.}
Another example is the two-dimensional incompressible Navier-Stokes equation. In the velocity form, without forcing,
\begin{equation*}
    \partial_t u(x,t) + u(x,t) \cdot \nabla u(x,t) =  -\nabla p(x, t) + \nu \Delta u(x,t),
\end{equation*}
The scaling is  by $u_\lambda(x,t) = u(\lambda x, \lambda t)$, $p_\lambda(x,t) = p(\lambda x, \lambda t)$, and $\nu_\lambda := \lambda^{-1} \nu$.
\sam{
In the vorticity formulation where $\omega = \mathrm{curl}(u)$, we do not need to rescale the time.
\begin{equation}
\label{eq:NS}
    \partial_t \omega(x,t) + u(x,t) \cdot \nabla \omega(x,t) = \nu \Delta \omega(x,t),
\end{equation}
Rescaling the spatial variable $x$ corresponds to rescaling  the viscosity $\nu$; $\omega_\lambda(x,t) = \omega(\lambda x, t)$ and $ u_\lambda(x,t) = \lambda^{-1}  u(\lambda x, t)$ solves 
\[
\partial_t \omega_\lambda(x,t) + u_\lambda(x,t) \cdot \nabla \omega_\lambda(x,t) = \nu_\lambda \Delta \omega_\lambda(x,t),
\]
where $\nu_\lambda := \lambda^{-2} \nu$, here the coefficient $\lambda$ in front of the term $(u_\lambda(x,t) \cdot \nabla \omega_\lambda(x,t))$ is absorbed by $u_\lambda$. 
}

%% file: Sections/4-Algorithm.tex
\subsection{Main algorithms}
Our main learning algorithm contains two parts: scaling down with data augmentation  and scaling up as self-supervised learning. We also discuss a test-time domain-decomposition scheme based on pre-trained scale-consistent model.

\textbf{Remark: neural operator automatically rescales input to unit length.} For standard neural networks such as convolution neural networks, re-scaling $\T$ needs to be implemented as interpolation. However, in the design of neural operators such as FNO, the domain size is implicitly re-scaled to unit length, where the Fourier basis is defined  with domain length $[0, 1]$. It means neural operators can directly work on various grid sizes generated from the sampling algorithm. Given $\T_\lambda f$ defined on domain $[0,\lambda]$, Fourier neural operator $\Psi$ automatically rescales it to unit length, 
\[
\Psi(\T_\lambda f, \ldots) := \Psi(\T_{1/\lambda} \T_\lambda f,\ldots)
 = \Psi(f,\ldots).
\]
where $f$ is defined on unit size $[0,1]$. Therefore, the re-scaling $\T$ is omitted in the algorithm.

\subsubsection{Scaling consistency loss for training}
In this section we discuss the scale consistency loss via up-sampling and downsampling. We sample various domain sizes with the same underlying resolution, which leads to new training instances with various grid size.

\paragraph{Sub-domain sampling.} The sub-domain sampling algorithm is based on equation \eqref{eq:aug}, where we use sub-sampling (i.e., restrict to sub-domain) to obtain instance with smaller scale $\lambda k < k$. Given the input and output data $\{(a, g, k), u\}$ defined on domain $\Omega$, we truncate the domain into a smaller sub-domain $\hat{\Omega}$. The input and output restriction to the sub-domain, along with the re-scaled parameter, become a new data instance $\{(\hat{a}, \hat{g}, \hat{k}), \hat{u}\}$. 
The new data instance share the same resolution as the original domain, and therefore smaller grid size. Therefore, no interpolation is required.
We compute the consistency loss as the difference between the model evaluated on restricted input $\Psi(\hat{a}, \hat{g}, \hat{k})$ and the restricted output $\hat{u}$.

\begin{algorithm}
\caption{Sub-domain sampling}
\label{algo:sub-sample}
\begin{algorithmic}[1]
\STATE \textbf{input}: data tuple of coefficient, boundary, scale parameter, and solution $\{(a, g, k), u\}$ on domain $\Omega = [0, 1]^2$, model $\Psi$, and sampling rate $\lambda  < 1$.
\STATE sample the sub-domain $\hat{\Omega} = [w, w+\lambda ]\times[h,h+\lambda]$, where $w,h \sim Unif[0, 1-\lambda]$.
\STATE define new instance \\
$(\hat{a} = a|_{\hat{\Omega}}, \hat{g} = u|_{\partial\hat{\Omega}}, \hat{k} = \lambda\ k), \hat{u} = u|_{\hat{\Omega}}$.
\STATE \textbf{output}: scale-consistency loss
$\|\Psi(\hat{a}, \hat{g}, \hat{k}) - \hat{u}\|$.
\end{algorithmic}
\end{algorithm}

\paragraph{Super-domain sampling.} The super-domain sampling algorithm is based on equation \eqref{eq:sc}, where we sample new instances corresponding to larger scale $\lambda k > k$. 
Given the distributions $\mu$ for $a$ and $\nu$ for $g$, we can sample new instance $a, g$ with larger scale $\lambda k$ and apply Algorithm \ref{algo:sub-sample}.
Different from \ref{algo:sub-sample}, we do not have the ground truth output $u$ on the larger scale. Instead, we estimate using the model  $u =  \Psi(a, g, \lambda k)$.

\begin{algorithm}
\caption{Training: super-domain sampling}
\label{algo:super-sample}
\begin{algorithmic}[1]
\STATE \textbf{input}: distributions of inputs coefficient and boundary $\mu, \nu$, model $\Psi$, scale parameter $k$, and sampling rate $\lambda  > 1$.
\STATE sample new instances $a\sim \mu, g\sim \nu$. \\
define new scale as $\lambda k$.
\STATE estimate the solution of new domain $u =  \Psi(a, g, \lambda k)$.
\STATE call Algo \ref{algo:sub-sample} with input
$\{(a, g, \lambda k), u\}$ and scale $1/\lambda$.
\STATE \textbf{output}: scale-consistency loss \\
$\| \Psi(a|_{\hat{\Omega}}, \Psi(a,g, \lambda k)|_{\partial\hat{\Omega}}, k) - \Psi(a,g,\lambda k)|_{\hat{\Omega}} \|$. 
\end{algorithmic}
\end{algorithm}


\subsubsection{Domain decomposition with pre-trained scale-consistent model}
Beyond the scale-consistency augmentation applied during training, we explore a test-time iterative domain decomposition (DD) \cite{huang2025operator, wang2022mosaic} refinement technique to further improve the performance of a pre-trained neural operator, particularly on large domains. This method leverages the operator's ability to solve smaller problems accurately by applying the neural operator multiple times on overlapping subdomains and iteratively merging these local solutions into an improved global solution. The approach trades additional computation for enhanced solution quality while maintaining coherent solutions across subdomain boundaries directly through the scale-consistency properties developed in this work. We demonstrate this technique for the Darcy flow problem.

The domain decomposition methodology, as described in Algorithm \ref{algo:domain-decomp}, decomposes a large domain $\Omega$ into overlapping subdomains $\{\Omega_i\}_{i=1}^N$ and iteratively refines the global solution by applying the neural operator to each subdomain with boundary conditions extracted from the current global solution estimate. For a 2D domain $\Omega = [0,1]^2$ discretized on a $s \times s$ grid, we partition it into a $4 \times 4$ array of overlapping patches, where each patch has size $(s/2  \times s/2)$ with $s/4$ overlap. The boundary conditions for each subdomain $\Omega_i$ are constructed by extracting values from the reference solution $u^{(k)}$ at iteration $k$, such that $g_i^{(k+1)} = u^{(k)}|_{\partial \Omega_i}$ for internal boundaries, while external boundaries use the true boundary conditions from the original problem.

\begin{algorithm}
\caption{Domain decomposition}
\label{algo:domain-decomp}
\begin{algorithmic}[1]
\STATE \textbf{input}: neural operator $\Psi$, coefficient field $a$, boundary condition $g$ on domain $\Omega$.
\STATE decompose $\Omega$ into overlapping patches $\{\Omega_i\}_{i=1}^N$ with overlap $\delta$.
\STATE extract boundary conditions for each patch: $g_i = u^{(k)}|_{\partial \Omega_i}$ where $u^{(k)}$ is current solution estimate.
\STATE solve local subproblems: $u_i = \Psi(a|_{\Omega_i}, g_i)$ for each patch $\Omega_i$.
\STATE \textbf{output}: merged global solution $u^{(k+1)} = \text{Blend}(\{u_i\}_{i=1}^N)$ using weighted averaging in overlap regions.
\end{algorithmic}
\end{algorithm}

To merge overlapping patch solutions, we employ a distance-based blending weight $w_i(x)$ that transitions smoothly from 1 at the patch center to 0 at the boundaries. The global solution is reconstructed as:
\begin{equation}
u^{(k+1)}(x) = \frac{\sum_{i: x \in \Omega_i} w_i(x) u_i(x)}{\sum_{i: x \in \Omega_i} w_i(x)}
\end{equation}

Compared to standard domain decomposition algorithms that initialize the interior boundary $g_i$ as zeros, the scale-consistent operator can initialize the interior with coarse prediction. Further, the domain decomposition scheme can be applied with multiple levels $\{s, s/2, s/2^2, \ldots\}$.

%% file: Sections/4-Method.tex
\section{Scale-Informed Neural Operator}\label{sec:model}
\begin{figure*}[t]
\begin{center}
\includegraphics[width=0.7\textwidth, height=0.3\textheight]{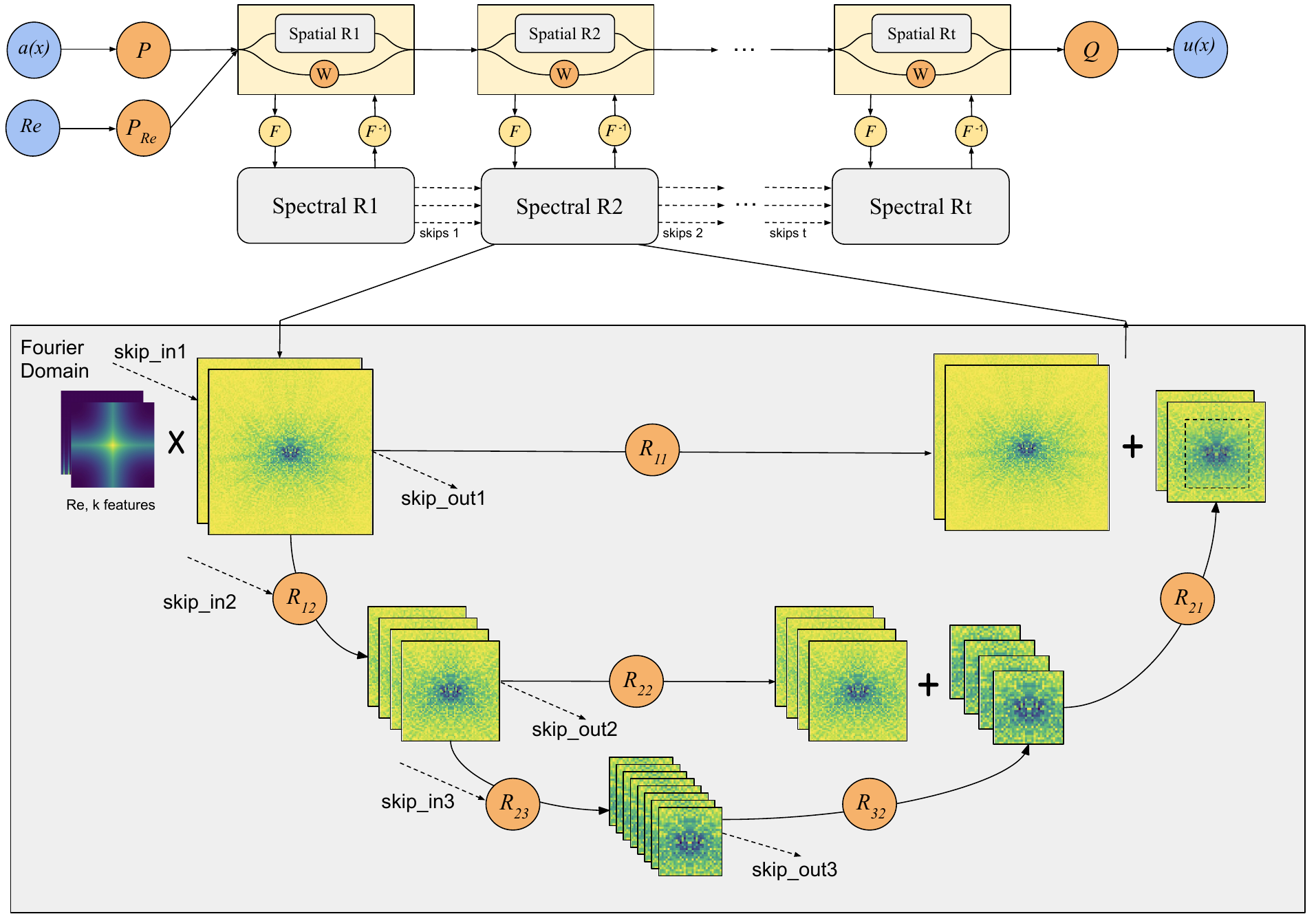}
\caption{The scale-informed neural operator has a U-shape structure on the Fourier space. The scale parameter (such as $Re$)  are embedded at each spectral layer. In the down block, the input tensors are truncated and lifted by complex layer $R$; in the up block, the tensors are projected and added to the inputs. Skip connections are added across the blocks. $P$ is the encoder and $Q$ is the decoder. Details in \ref{sec:ushape}.
}
\label{fig:architecture}
\end{center}
\end{figure*}

The scale-informed neural operator is based on the FNO \cite{li2020fourier}, where convolution is implemented as a pointwise multiplication in the Fourier space. Since FNO automatically rescales its input to unit length, we design a scale embedding in the Fourier space to inform the model of the scale parameter $k$. Furthermore, we design a U-shaped architecture to optimize the channel dimension.

\subsection{Embed scale parameters in  Fourier Space}

In the previous FNO, the weight tensor $R$ is defined as a $(M_1 \times \cdots \times M_d \times C_{in} \times C_{out})$-tensor, which is sufficient for lower-dimensional problems with fewer total modes $M$. For larger-scale problems, such as highly turbulent flows, the weight tensor $R$ becomes prohibitively large. Therefore, we propose an implicit representation of the weight tensor similar to AFNO \cite{AFNO}, where the complex weight $R$ with the shape $(C_{in} \times C_{out})$ is shared across all modes $(M_1 \times \cdots \times M_d)$. 

Different from AFNO, we further define the features of scale $k$ and mode index $\xi$ as input, so that the transform $R$ can behave correspondingly with respect to different scales $k$ and modes $\xi$. Let $C$ be the embedding channel dimension; we define scale features as $h(k)_i =  k^{i/(C-1)}$ for $i = 0,1,\ldots,C-1$, which covers a wide range from $k^{0/(C-1)}=1$ to $k^{(C-1)/(C-1)}=k$. The input $f_t(\xi) \in \mathbb{C}^{C_{in}}$ is first element-wise multiplied with the features of the scale parameter and wavenumber $h(k ,\xi)$, and then multiplied with $R$, followed by a group normalization and a complex activation $\sigma$ as defined in Section \ref{sec:complex_activation}. The transform $\mathcal{K}$ can be viewed as a kernel function defined on the Fourier space: 
\begin{equation}
\label{eq:embedding}
    (\mathcal{K}f_{t+1})(\xi) = \sigma \bigl(R (f_t(\xi) \odot h(k ,\xi))\bigr).
\end{equation}

\subsection{Multi-band Architecture}
The Fourier signal usually follows an ordered structure, where the energy decays as the wavenumber increases. Therefore, previous methods such as FNO \cite{li2020fourier} and SNO \cite{fanaskov2022spectral} choose to truncate to a fixed number of frequencies by omitting higher frequencies. Similar to previous works such as UNet \cite{ronneberger2015u}, UNO\cite{uno}, and multi-wavelet operator \cite{gupta2021multiwavelet}, we design a multi-band structure to gradually shrink the frequency bands, as shown in Figure \ref{fig:architecture}. Different from UNO, which applies spectral convolutions at each down and up block, in this work, we define the U-shaped structure fully in the Fourier space. Given the initial channel dimension $C$, maximum input modes $M$, and a predefined number of levels $L$, we define $C_l$ and $M_l$ as $C_l = 2^l C$ and $M_l = 2^{-l} M$, where each block has shape $C_l^2 M_l^d$. For $d=2$, $C_l^2 M_l^2 = C^2 M^2$, so each level has the same size. We define the first level using the weight-sharing formulation, where $R_1$ has the shape $(C_{in} \times C_{out})$, and higher levels in tensor formulation with $(M_l^d \times C_{in} \times C_{out})$. The detailed implementation is described in \ref{sec:ushape}.

\subsection{Boundary condition}
For boundary value problems, we take the boundary as an additional input. For a 1-dimensional boundary on a 2-dimensional square domain, we extend the boundary to 2D by repeating along the other dimension. For Dirichlet-type boundaries, it is known that the boundary is the restriction of the solution, and their magnitudes should be similar. Therefore, we define a normalization at the end of the model that multiplies the output by the magnitude of the boundary.

%% file: Sections/5-Experiments.tex
\section{Experiments}

\begin{table}[t]
\centering
\caption{Comparison of FNO, UNet, UNO, CNO, and SINO with and without scale-consistency. 
Models are trained at certain scale and zero-shot test across others. 
Overall, scale-consistency helps each model extrapolate to unseen scales. 
Errors are in relative-L2 (\num{e-2}). The Darcy Flow is scale-invariant so the SINO does not apply.}
\label{table:darcy_burgers}
\begin{tabular}{l|cccccc}
\toprule
\midrule
\multicolumn{7}{l}{\textbf{Darcy Flow (scale of coefficient functions)}} \\
\midrule
Model & 2 & 3 & 4 (training) & 8 & 16 & \\
\midrule
FNO & 3.921 & 3.842 & 3.737 & 3.323 & 3.214 & \\
FNO+scale & \textbf{1.990} & \textbf{1.932} & \textbf{1.990} & \textbf{2.130} & \textbf{2.300} & \\
UNet & 6.638 & 6.981 & 6.011 & 5.527 & 6.361 & \\
UNet+scale & 5.130 & 4.945 & 5.869 & 5.645 & 6.094 & \\
UNO & 5.534 & 5.336 & 4.725 & 4.495 & 4.366 & \\
UNO+scale & 3.009 & 2.753 & 3.087 & 4.602 & 4.600 & \\
CNO & 4.393 & 4.281 & 4.248 & 4.159 & 4.451 & \\
CNO+scale & 4.149 & 4.228 & 4.505 & 4.418 & 4.716 & \\
\midrule
\midrule
\multicolumn{7}{l}{\textbf{Burgers' Equation (Viscosity $\nu$)}} \\
\midrule
Model & 1/100 & 1/200 & 1/400 (training) & 1/1000 & & \\
\midrule
FNO & 28.602 & 11.005 & 1.230 & 8.709 & & \\
FNO+scale & 27.799 & 10.008 & 1.908 & 9.442 & & \\
SINO & 24.914 & 10.027 & 1.174 & 8.363 & & \\
SINO+scale & \textbf{5.926} & \textbf{1.720} & \textbf{0.957} & \textbf{4.575} & & \\
UNet & 32.897 & 22.463 & 20.119 & 26.481 & & \\
UNet+scale & 30.137 & 22.815 & 25.138 & 30.747 & & \\
UNO & 28.581 & 10.963 & 1.235 & 8.624 & & \\
UNO+scale & 28.716 & 11.009 & 1.387 & 8.720 & & \\
CNO & 27.999 & 10.461 & 2.059 & 9.054 & & \\
CNO+scale & 25.264 & 8.280 & 3.959 & 11.191 & & \\
\midrule
\bottomrule
\end{tabular}
\end{table}

We generated datasets for the Darcy Flow, Helmholtz equation, and Navier-Stokes equation, each spanning a wide range of scales. For each test case, we trained the models on a narrow range of scales and compared the performance with and without self-consistency augmentation. All experiments were run on Nvidia A100 (80GB, 40GB) or P100 (16GB) GPUs. 
The error metric is relative L2 error. The choice of hyperparameters can be found in Appendix \ref{app:exp}. The results show that self-consistency augmentation helps the model generalize better to unseen scales.

\subsection{Self-consistency loss for training scale-consistent neural operator.}
In the first part, we compare FNO, UNet, and our models, with and without the self-consistency loss. For Darcy and Helmholtz equations, where the input distribution is given as a Gaussian random field, we apply both sub-sampling \ref{algo:sub-sample} and super-sampling \ref{algo:super-sample}. For the Navier-Stokes equation, the input distribution is unknown, so we only apply sub-sampling. The detailed data generation can be found at \ref{sec:data}.

\textbf{Darcy Flow.}
We considered the Darcy Flow \eqref{eq:darcy} with a non-zero Dirichlet boundary. The input coefficient is sampled at different scale, as described in \ref{sec:data}.
The resolutions were $s = {64, 96, 128, 256, 512}$, respectively. We train 1024 instances for training and 128 for testing. The data generation details can be found in Appendix \ref{sec:data-darcy}. We used $\sigma = 1$ for training. Since Darcy has no scale parameters, we used FNO with and without scale-consistency. As shown in Table \ref{table:darcy_burgers}, FNO with scale-consistency reduced the error by half compared to the baseline.

\textbf{Helmholtz Equation.}
We considered the Helmholtz equation \eqref{eq:helmholtz} with a non-zero Dirichlet boundary. The input coefficients $a, g$ were sampled from a fixed Gaussian random field, with varying wavenumbers $k = {1, 2, 5, 10, 25, 50, 100}$. The resolutions were ${64, 64, 64, 128, 256, 512, 1024}$, respectively. We train 1024 instances for training and 128 for testing. The data generation details can be found in Appendix \ref{sec:data-helm}. We used $k ={5, 10, 25}$ for training. The scale-informed neural operator with scale-consistency reduced the error by half compared to the baseline FNO on smaller wavenumbers $k={1, 2}$, but neither model captured larger scales $k={50, 100}$, since Helmholtz equation has very different behaviors on larger scales.

\begin{table}[t]
\centering
\caption{Comparison of FNO, UNet, UNO, and SINO with and without scale-consistency, continued. 
Models are trained at certain scale and zero-shot test across others. 
Overall, scale-consistency helps each model extrapolate to unseen scales. 
Errors are in relative-L2 (\num{e-2}).}
\label{table:helm_ns}
\begin{tabular}{l|cccccc}
\toprule
\midrule
\multicolumn{7}{l}{\textbf{Helmholtz Equation (Wave number $k$)}} \\
\midrule
Model & 1 & 2 & 5 (tr) & 10 (tr) & 25 (tr) & 50 \\
\midrule
FNO & 136.847 & 131.200 & 4.285 & 12.575 & 21.060 & 107.186 \\
FNO+scale & 44.625 & 36.026 & 3.186 & \textbf{11.924} & 19.744 & 108.916 \\
SINO-U & 69.437 & 63.283 & 3.666 & 12.503 & 19.728 & \textbf{102.980} \\
SINO-U+scale & \textbf{8.960} & \textbf{6.960} & \textbf{3.081} & 12.490 & \textbf{19.001} & 112.940 \\
UNet & 164.945 & 156.775 & 48.341 & 51.028 & 21.189 & 112.914 \\
UNet+scale & 51.441 & 64.313 & 63.827 & 52.731 & 53.541 & 104.477 \\
UNO & 120.742 & 101.478 & 9.350 & 16.172 & 32.280 & 118.570 \\
UNO+scale & 125.742 & 91.541 & 10.821 & 19.605 & 36.017 & 117.776 \\
\midrule
\midrule
\multicolumn{7}{l}{\textbf{Navier-Stokes (Reynolds number Re)}} \\
\midrule
Model & 250 & 500 & 1000 (training) & 2000 & 4000 & 10000 \\
\midrule
FNO & 0.447 & 0.750 & 1.015 & 3.108 & 7.374 & 18.295 \\
FNO+scale & \textbf{0.302} & 0.531 & \textbf{0.743} & 2.446 & 6.137 & 17.127 \\
SINO-U & 0.695 & 0.782 & 0.976 & 2.466 & 4.793 & 13.772 \\
SINO-U+scale & 0.357 & \textbf{0.514} & 0.953 & 2.186 & \textbf{4.289} & \textbf{11.483} \\
UNet & 4.156 & 2.706 & 0.809 & \textbf{2.096} & 10.027 & 22.284 \\
UNet+scale & 1.086 & 1.753 & 13.802 & 15.427 & 16.442 & 28.297 \\
UNO & 4.228 & 3.021 & 4.147 & 8.316 & 16.728 & 33.221 \\
UNO+scale & 4.005 & 2.661 & 3.458 & 6.941 & 14.785 & 30.663 \\
\bottomrule
\end{tabular}
\end{table}

\textbf{Burgers' Equation.}
We considered the Burgers' equation \eqref{eq:burgers}. Given the initial condition and time-dependent boundary condition as input, the model predicts the solution over the next time interval. We train the FNO model and the multi-scale neural operator model (ours) with and without scale-consistency loss. The scale-consistent loss is across both the spatial and temporal domain. The models are trained on viscosity $=1/400$ and tested on viscosities $\nu=1/100, 1/200, 1/400, 1/1000$.
The multi-scale neural operator with scale-consistency reduced the error up to 5x compared to the baseline FNO on unseen viscosity.

\textbf{Navier-Stokes Equation (autoregressive).}
We considered the Navier-Stokes equation \eqref{eq:NS} defined on sub-domain similar to applications in climate. The input is the vorticity field of the previous ten time frames $\omega_0$. We considered Reynolds numbers ranging from $Re = {250, 500, 1000, 2000, 4000, 10000}$. The resolutions were ${32, 64, 128, 128, 256, 512}$, respectively. We train 50 trajectories for training and 5 (per each $Re$) for testing, where each trajectory consists of 300 time steps, with $dt=0.1$. The data generation details can be found in Appendix \ref{sec:data-NS}. We used $Re ={1000}$ for training. The multi-scale multi-band neural operator with scale-consistency reduced the error by 1/4 compared to the baseline UNet on unseen $Re ={250, 500, 4000, 10000}$.

\textbf{Navier-Stokes Equation (space-time, 2+1 dimensional).}
We also considered the spatiotemporal modeling for the Navier Stokes equation, velocity formulation. Similar to the autoregressive setting above, we considered Reynolds numbers ranging from $Re = {250, 500, 1000, 2000, 4000, 10000}$. For continuous-time modeling, we use $dt=1/256$ and are given input of the history, consisting of 24 frames, to predict the next 24 frames. In Table~\ref{table:NS-spacetime}, we observe significant improvements with scale embedding and spatiotemporal cropping for out-of-distribution Reynolds numbers. Improvements are highlighted in Table \ref{table:NS-spacetime}.

\textbf{Comparison with symmetry-based augmentation.}
On the Darcy flow, we compare the scale-consistency augmentation with existing symmetry-based augmentation as used in \cite{wang2020incorporating, brandstetter2022lie}. As shown in Table \ref{table:symmetry}, scale-consistency augmentation leads to better generalization compared to rotation plus reflection. Furthermore, scale-consistency works seamlessly with rotation and reflection. The best result is achieved by combining the three augmentation methods together.

\begin{table}[t]
\centering
\caption{Comparison of scale-consistency with existing symmetries for data augmentation, in relative-L2 error (\num{e-2}). We train FNO  on Darcy flow at $scale=4$ and zero-shot test at other scales.}
\label{table:symmetry}
\begin{tabular}{l|ccccc}
\toprule
Scale & 2 & 3 & 4 & 8 & 16 \\
\midrule
No aug. & 4.143 & 4.193 & 4.036 & 3.552 & 3.352 \\
Rot. & 3.101 & 2.944 & 2.953 & 2.797 & 2.872 \\
Ref. & 2.821 & 2.701 & 2.597 & 2.616 & 2.684 \\
Rot.+Ref. & 2.713 & 2.469 & 2.450 & 2.461 & 2.582 \\
\midrule
Scale (ours) & 1.918 & 2.075 & 2.035 & 2.159 & \textbf{2.237} \\
All (ours) & \textbf{1.903} & \textbf{1.816} & \textbf{1.910} & \textbf{2.095} & 2.309 \\
\bottomrule
\end{tabular}
\end{table}

\subsection{Test-time correction with domain decomposition}

\begin{figure}[t]
    \centering
    \includegraphics[width=\linewidth]{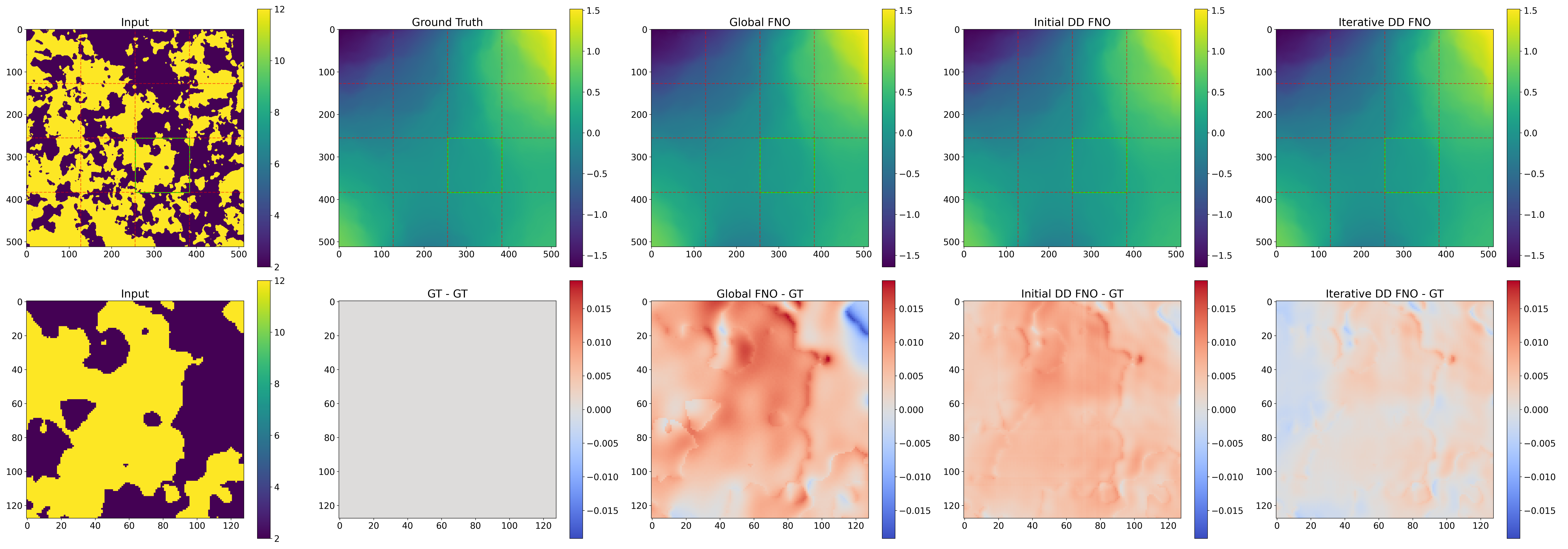}
    \caption{Domain decomposition with pre-trained scale-consistent neural operators. The global domain (top) is  decomposed into $16$ overlapping sub-domains (bottom). The subdomains are iteratively refined by the same pre-trained operator.}
    \label{fig:dd}
\end{figure}

We apply the iterative domain-decomposition refinement  algorithm \ref{algo:domain-decomp} to a pre-trained FNO model (trained on a dataset of $128 \times 128$ (scale $= 4$) with scale-consistency loss) for the 2D Darcy flow problem on a $512 \times 512$ resolution target (scale $= 3$), as in Figure \ref{fig:dd}. The domain is decomposed into $16$ sub-domains with overlaps. Each inner sub-domain is $128 \times 128$, and we use a fixed overlap of $128$ pixels, resulting in effective outer sub-domain sizes of $256 \times 256$ for processing by the FNO. We use blending masks with a linear ramp over $16$ pixels at the edges of the overlap.

The performance is summarized in Table \ref{tab:dd}. Iteration 0 represents the standard FNO applied to the full domain. Iteration 1 is the first pass of the DD method, using $u^{(0)}$ for internal boundaries. Subsequent iterations refine this solution. The method achieves a 40\% reduction in relative L2 error compared to direct application on the global domain, with convergence typically occurring within a couple of iterations. The optimal overlap parameter is problem-dependent but generally ranges from 50\% to 100\% of the patch size.

\begin{table}[h]
\centering
\caption{Relative L2 error reduction on Darcy Flow ($512 \times 512$) using iterative domain decomposition refinement. Overlap was fixed at $128$ pixels.}
\label{tab:dd}
\begin{tabular}{l|cc}
\toprule
Iteration $k$ & Avg. Relative L2 Error (\num{e-2}) & Improvement over Global FNO (\%) \\
\midrule
0 (Global FNO)       & 5.3162 & N/A \\
1 (Init. DD FNO)     & 3.4571 & 34.97 \\
2 (Iter. DD FNO)     & 3.2464 & 38.93 \\
3 (Iter. DD FNO)     & 3.2029 & 39.75 \\
4 (Iter. DD FNO)     & 3.1868 & 40.06 \\
5 (Iter. DD FNO)     & \textbf{3.1801} & \textbf{40.18} \\
\bottomrule
\end{tabular}
\end{table}

The efficacy of the iterative DD approach as a method to scale test-time computation and increase the accuracy of pre-trained neural operators on large-scale problems by enforcing a form of local consistency and iteratively propagating this information is promising.

\subsection{Ablation studies on the model architecture}
\textbf{Embed scale parameter in the frequency space.}
We conducted several ablation studies on the proposed model architecture along with scale-consistency loss. We test the scale embedding and positional embedding on the frequency space \eqref{eq:embedding} with Burgers' equation, Helmholtz equation, and Navier-Stokes equation. As shown in Table \ref{table:ScaleFrequency} and \ref{table:NS-spacetime}, the embedding in general improves the performance. We sometimes find the scale parameter is unnecessary in the Navier-Stokes equation when it can be inferred from the history of trajectory.

\begin{table}
    \centering
    \caption{Ablation for scale-informed neural operator on different equations in relative-L2 error (\num{e-2}). For Burgers' equation, we train on viscosity $\nu=1/400$ and zero-shot test on other scales. For Helmholtz equation, we train on wavenumber $k=5, 10, 25$.}
    \label{table:ScaleFrequency}
    \begin{tabular}{cc|cccccc}
        \toprule
        \multicolumn{8}{c}{Burgers' equation} \\
        \midrule
        \textbf{Scale Informed} & \textbf{Freq. Emb.} & $\nu$ = 1/100 & $\nu$ = 1/200 & $\nu$ = 1/400 & $\nu$ = 1/800 & & \\
        \midrule
        No & No & 28.531 & 10.756 & 1.087 & 8.889 & --- & --- \\
        No & Yes & 28.660 & 10.832 & \textbf{0.916} & 8.725 & --- & --- \\
        Yes & No & 10.731 & 2.540 & 1.055 & 5.477 & --- & --- \\
        Yes & Yes & \textbf{6.334} & \textbf{1.887} & 1.042 & \textbf{4.636} & --- & --- \\
        \midrule
        \multicolumn{8}{c}{Helmholtz equation} \\
        \midrule
        \textbf{Scale Informed} & \textbf{Freq. Emb.} & k = 1 & k = 2 & k = 5 & k = 10 & k = 25 & k = 50 \\
        \midrule
        No & No & 17.914 & 5.963 & 3.537 & 11.042 & 16.338 & 106.597 \\
        No & Yes & 15.642 & 6.384 & 3.441 & 10.294 & 14.056 & 102.015 \\
        Yes & No & 16.741 & 4.822 & 2.914 & 10.631 & 12.989 & 103.151 \\
        Yes & Yes & \textbf{9.438} & \textbf{4.980} & \textbf{2.921} & \textbf{9.874} & \textbf{11.574} & \textbf{93.938} \\
        \bottomrule
    \end{tabular}
\end{table}

\begin{figure*}[t]
\begin{center}
\includegraphics[width=0.45\columnwidth]{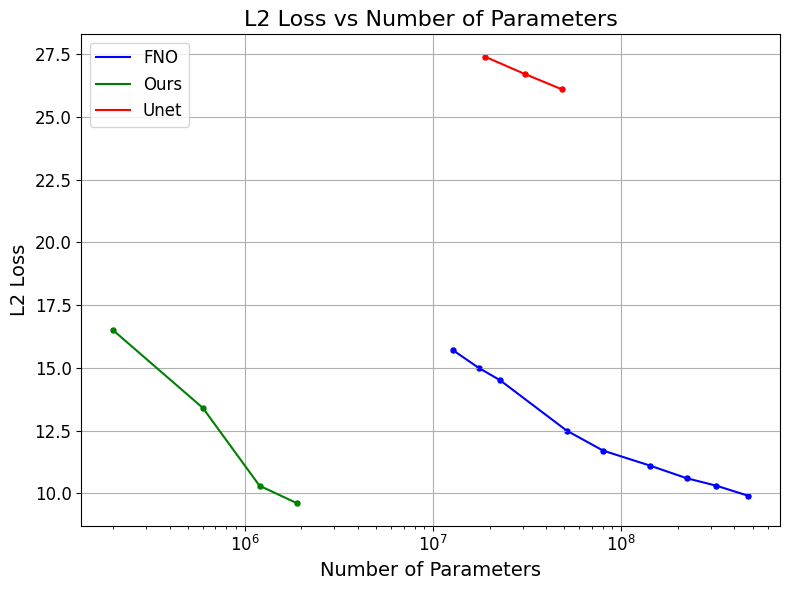} 
\includegraphics[width=0.5\columnwidth]{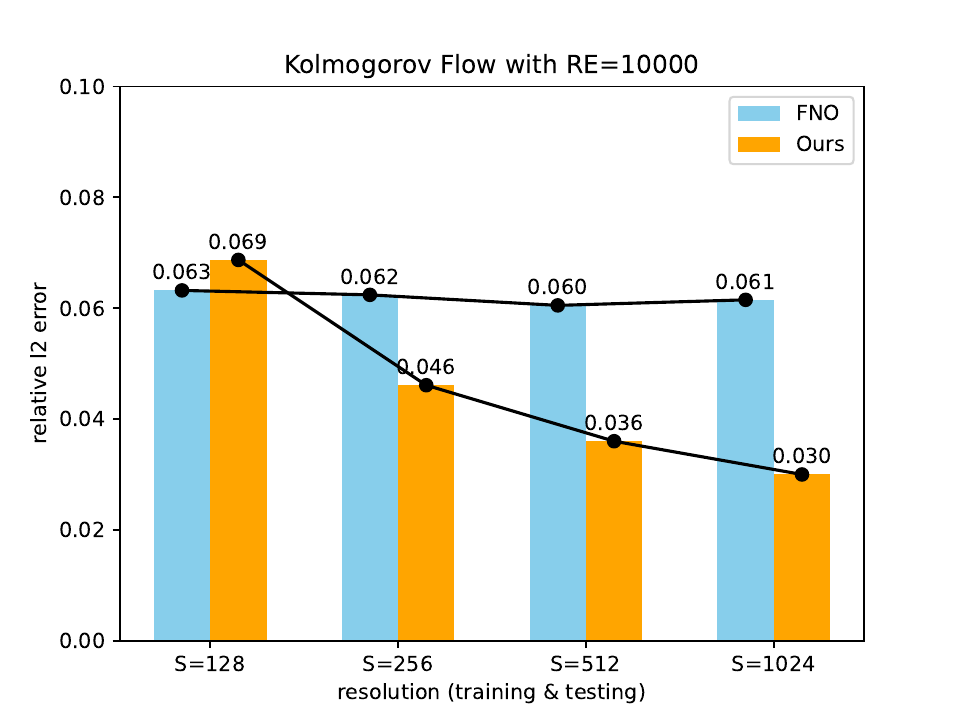} 
\caption{Ablation study. \textbf{left}: Cost-Accuracy: we train and test each model at various sizes on Kolmogorov Flow with RE=5000.  Our model (u-shape)  converges faster than baseline models. Further, the model (shared) achieves comparative accuracy with 1/10 of the parameters.
\textbf{right}: discretization convergence:
the proposed model does not truncate to a fixed bandwidth. As the training resolution increases, the model's error converges while the baseline FNO remains the same.
}
\label{fig:ablation}
\end{center}
\end{figure*}

\textbf{U-shape structure and shared kernel.} 
We further conducted ablation studies on the U-shape structure model in the standard supervised learning setting on periodic Navier-Stokes equation with fixed scales $Re=5000$ (with forcing) and $Re=10000$ (zero forcing) as in \cite{li2022learning}. 
For baselines, we consider FNO \cite{li2020fourier}, UNet \cite{ronneberger2015u}, FNO-UNet \cite{gupta2022towards}, and UNO \cite{uno}.
The results show that our model achieves a smaller error rate with one-tenth of the parameters compared to the previous FNO at the cost of longer runtime, as shown in Figure \ref{fig:ablation} (left). Since the model does not truncate the maximum Fourier frequency, its accuracy improves as the resolution refines, as shown in Figure \ref{fig:ablation} (bottom right).

%% file: Sections/7-Conclusion.tex
\section{Conclusion}
In this paper, we consider the scale consistency for learning solution operators on PDEs across various scales. By leveraging the scale-consistency properties of PDEs and designing a scale-informed neural operator, we demonstrated the ability to model a wide range of scales.  Experimental results showed significant improvements in generalization to unseen scales, with better generalization errors compared to baseline models.
This approach holds promise for improving the efficiency and generalizability of data-driven PDE solvers, reducing the need for extensive training data, and enabling the development of more flexible and foundational models for scientific and engineering applications.

\textbf{Limitations and future work.}
In this work, we make an assumption of the system is governed by a set of partial differential equations with changing scales. Such assumption makes it possible to generalize and extrapolate to the behaviors at the unseen scales. In practice, sometimes the micro-scale physics cannot be described by the same set of PDEs. For example, the molecular dynamics cannot be generalized from the continuum model. In this case, additional micro-scale data and equations will be required to fine-tune the model.

While sub-sampling (Algorithm \ref{algo:sub-sample}) is generally helpful, super-sampling (Algorithm \ref{algo:super-sample}) requires input distribution known to sample new instances. While the super-sampling works well for Darcy and Burgers, it is challenging to subsample from the attractor for the Navier-Stokes equation. As a potential future direction, it could be an interesting direction to combine with generative models \cite{lim2023score} to sample virtual inputs. 


%% file: Sections/Acknowledgements.tex
\section*{Acknowledgements}
Anima Anandkumar is supported by the Bren named chair professorship, Schmidt AI2050 senior fellowship, and ONR (MURI grant N00014-18-1-2624).

%% file: Sections/A-Data.tex
\section{Datasets}
\label{sec:data}


\subsection{Darcy Flow}
\label{sec:data-darcy}

We use a finite element solver with a resolution of 1024 to generate the dataset. The dataset is similar to the one used in \cite{li2020fourier}, but with non-zero Dirichlet boundary conditions.

\sam{
The input coefficient $a$ was sampled as $a = 2 + 10\cdot \mathds{1}_{[\hat{a}>0]}$ representing two types of media with values $2$ and $10$, where $\hat{a}$ is sampled from a Gaussian random field $\mathcal{N}(0,\mathcal{C})$. The covariance kernel $\mathcal{C}$ has Fourier coefficients $\exp(-\sigma |\xi|^{1/2})$. We considered wave lengths $\sigma = {2, 4/3, 1, 1/2, 1/4} := 1/k$, which is inverse to the scales.
}

\subsection{Helmholtz Equation}
\label{sec:data-helm}
We consider inhomogeneous Helmholtz Equation \cite{chen2023exponentially} in the form of 
\begin{equation}
\label{eqn:Helmholtz smooth k}
\left\{
\begin{aligned}
-\nabla \cdot(a\nabla u)-k^{2} u&=f, \ \text{in} \ \Omega\\
a\nabla u\cdot\nu&=ik\beta u+g, \ \text{on} \ \partial \Omega \, .
\end{aligned}
\right.
\end{equation}
Here, $a$ is the coefficient. $g$ is the boundary data. The forcing $f$ is fixed. We choose $\Omega=[0,1]^2$.

We prepare the data using a finite element solver with a resolution of 1024. It is worth noting that for physical equations, the Helmholtz equation is often paired with an impedance boundary condition, namely a Robin boundary condition:

$$\nabla u\cdot\nu=ik\beta u+g$$

\begin{figure}
    \centering
    \includegraphics[width=0.9\linewidth]{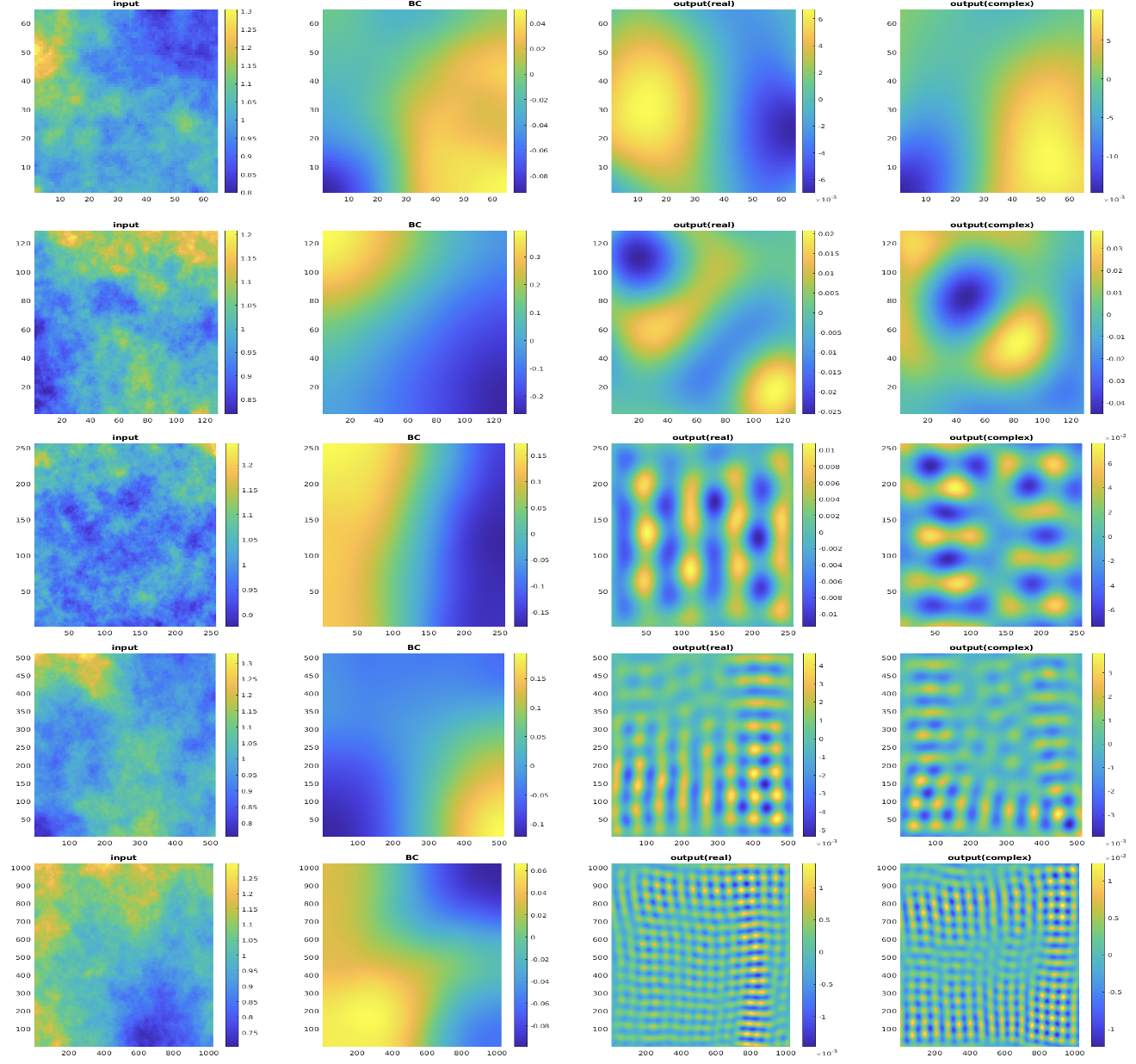}
    \caption{Helmholtz equations at multiple scales (wavenumbers). The five rows correspond to wavenumbers of $5,10,25,50,100$. The first column is the coefficient $a$; the second column is the boundary condition $g$; the third column is the corresponding solutions (real part); the fourth column is the solution (imaginary part).}
    \label{fig:helmholtz-data}
\end{figure}

For simplicity, we use Dirichlet boundary conditions for operator learning in this work. It is important to note that the Helmholtz equation with Dirichlet boundary conditions is a wave scattering problem, which may have multiple solutions as studied in \cite{de2022cost}. The Helmholtz equation dataset is visualized in Figure \ref{fig:helmholtz-data}.

\subsection{Burgers Equation}
\label{sec:data-burger}
We consider the Burgers equation in viscous form, similar to the setting in \cite{li2020fourier},
\begin{equation}
\label{eq:burgers}
    \partial_t u(x,t) + \partial_x (u^2(x,t)/2) = \nu \partial_{xx} u(x,t),
\end{equation}
Here we treat the time variable $t$ similar to the spatial variable $x$ Rescaling the spatial variable $x$ and temporal variable $t$ corresponding to rescaling the viscosity $\nu$ by $u_\lambda(x,t) = u(\lambda x, \lambda t)$ as shown in Figure \ref{fig:burgers}. By scaling $x$ and $t$ simultaneously, we balance the coefficient of $\partial_t u(x,t)$ and $ \partial_x (u^2(x,t)/2)$.
\begin{equation}
    \partial_t u_\lambda(x,t) + \partial_x (u_\lambda ^2(x,t)/2) = \nu_\lambda \partial_{xx} u_\lambda(x,t),
\end{equation}
where $\nu_\lambda = \lambda^{-1} \nu$

We consider a spatial-temporal formulation with resolution $256 \times 100$. We use a pseodu-spectral solver to general the dataset.

\subsection{Navier-Stokes Equation}
\label{sec:data-NS}

We consider a partially observed Navier-Stokes equation, inspired by practical applications in weather forecasting and oceanography, where a specific subdomain of the globe is of interest. We generate an isotropic Navier-Stokes equation on a periodic domain $[0, 1]^2$ and truncate it to a $[0, 0.5] \times [0, 0.5]$ subdomain. For convenience, we set the forcing term to zero and study the decay of turbulence.

\begin{figure}
    \centering
    \includegraphics[width=0.9\linewidth]{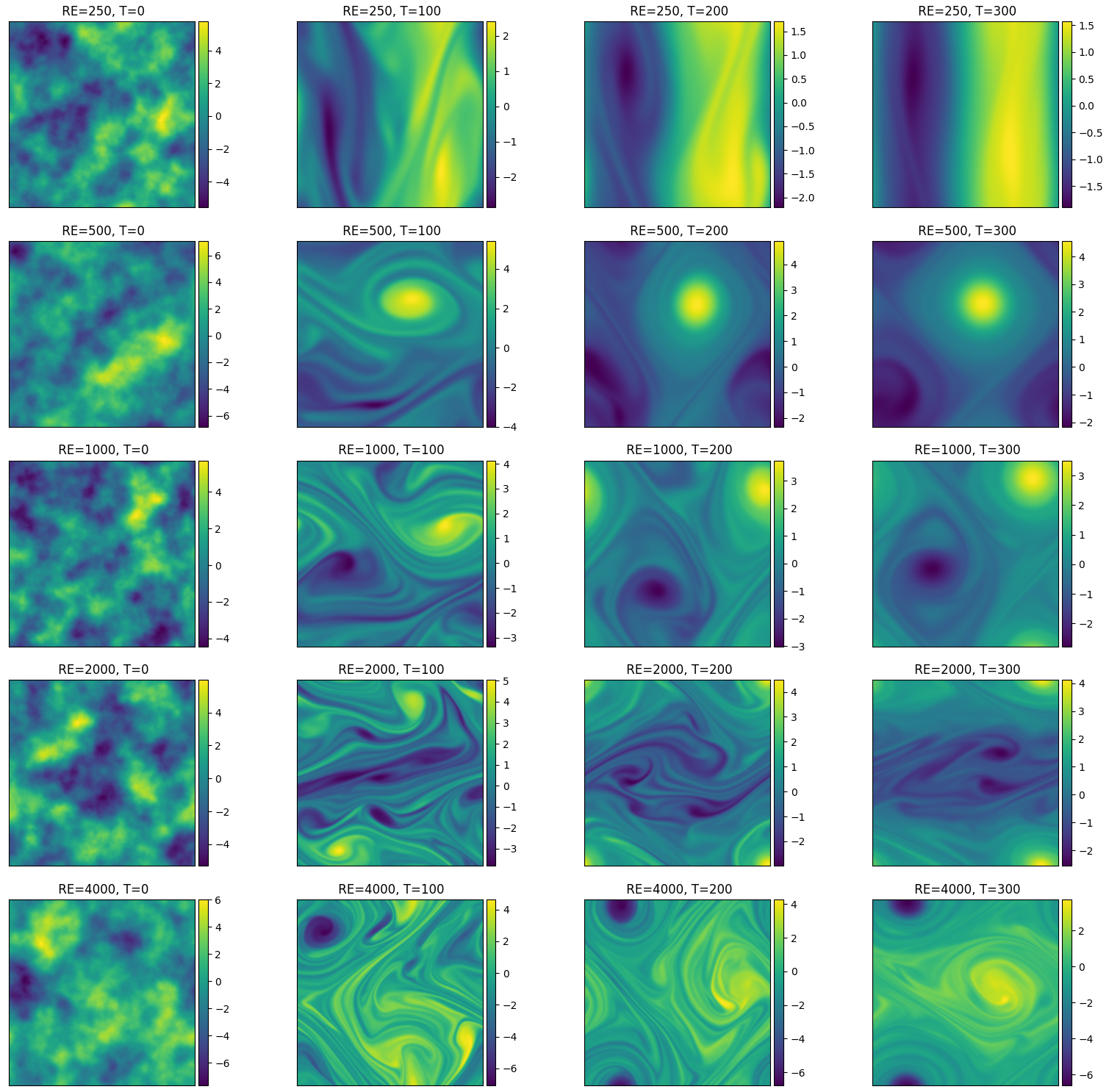}
    \caption{Navier-Stokes equations at multiple scales (Reynolds numbers). Rows correspond to scale and columns correspond to time steps.}
    \label{fig:ns-data}
\end{figure}

Since the underlying system is defined on a periodic boundary, we generate the data using a pseudo-spectral solver with Crank-Nicolson time updates. The Navier-Stokes equation dataset is visualized in Figure \ref{fig:ns-data}. We consider two different formulations, the auto-regressive formulation of coarse time step $dt=0.1$ and a continuous-time (2+1) formulation with a finer time step $dt=1/256$.

%% file: Sections/A-Theory.tex
\section{Proof of Theorem \ref{thm:sc}}

\subsection{Self-consistency loss}

Many scientific models are expressed via a partial differential equation (PDE). Fundamentally, a PDE expresses our (physics-)knowledge about correlations of solutions of this PDE, at infinitesimal length- and time-scales. Solving the PDE can therefore be thought of as the task of generalizing from these known infinitesimal correlations to macroscopic correlations.

Such PDEs often have well-defined scaling properties, in the sense that re-scaling one solution of the PDE formally gives rise to another solution of either the exact same PDE or of a PDE with rescaled coefficients and coefficient fields. 

Let $\cG: (a,g) \mapsto u=\cG(a,g)$ be the solution operator associated with a general PDE, consisting of a differential loss $\cP = \cP(u;a)$ and a boundary condition $\cB = \cB(u;g)$:
\[
\left\{
\begin{aligned}
\cP(u;a) &= 0, \quad \text{in }\Omega, \\
\cB(u;g) &= 0, \quad \text{on }\partial\Omega.
\end{aligned}
\right.
\]

\paragraph{Darcy flow.}
Here the PDE residual is $\cP(u;a) = -\nabla \cdot (a \nabla u)$, and the boundary condition $\cB(u;g) = u - g$. The solution operator $\cG: (a,g) \mapsto u = \cG(a,g)$ satisfies 
\[
\tau_{\Omega'} \cG(a,g) = \cG(\tau_{\Omega'}a, \tau_{\partial \Omega'} u).
\]

\paragraph{Helmholtz equation.}
Here the PDE residual is $\cP(u;f,k) = \Delta u + k^2 u - f$, and the boundary condition $\cB(u;g) = u - g$ (Dirichlet BC). The solution operator $\cG: (f,k,g) \mapsto u = \cG(f,k,g)$ satisfies 
\[
\tau_{\Omega'} \cG(f,k,g) = \cG(\tau_{\Omega'}f, \tau_{\Omega'}k, \tau_{\partial \Omega'} u).
\]

\subsection{(Exact) self-consistency implies generalization.}

How does self-consistency allow a neural operator $\Psi$ to generalize beyond the training data? To derive a corresponding mathematical result, we consider the case of the elliptic Darcy flow PDE. Here, the underlying solution operator $\cG$ maps $\cG: (a,g) \mapsto u$, where $a$ is the coefficient field, $g$ is the (Dirichlet) boundary condition and $u = \cG(a,g)$ is the solution of the following PDE:
\begin{align*}
\left\{
\begin{aligned}
-\nabla \cdot (a\nabla u) &= 0,  \quad (\text{in } \Omega),\\
u &= g, \quad (\text{on } \partial \Omega).
\end{aligned}
\right.
\end{align*}
We fix $\Omega = [0,1]^d$, and we assume throughout that all considered $(a,g)$ satisfy
\begin{align}
\label{eq:coercive}
0 < \lambda \le a(x) \le \Lambda, 
\quad 
\Vert g \Vert_{L^\infty(\partial \Omega)} \le 1, 
\quad 
a\in C^1(\Omega), \; g \in C^1(\partial \Omega).
\end{align}
The bounds on $a(x)$ represent a (uniform) coercivity condition, which is required to guarantee the well-posedness of the elliptic PDE, and therefore natural. The second bound on $g$ can essentially be made without loss of generality, since the elliptic PDE is linear in $g$, and hence $g$-inputs can usually be normalized to guarantee this constraint. The additional smoothness assumption on $a$ and $g$ (continuously differentiable) is made for simplicity, and could be considerably weakened at the expense of making the argument much more mathematically involved. We will only consider the simpler case \eqref{eq:coercive} here.

\label{sec:proof}
The following informal result summarizes our main theoretical insight and relevant conditions, without being overloaded with mathematical notation. 
\begin{theorem}
\label{thm:sc00}
Suppose that the neural network solution operator $\Psi$ is scale-consistent and is accurate for near-constant inputs. Namely if 
\begin{enumerate}
\item For almost constants $a$, we have
\[ 
\Psi(a,g) = \cG(a,g),
\] 
\item $\Psi$ satisfies  \eqref{eq:scaling} exactly along with translation symmetry,
\item $\Psi$ satisfies the boundary condition exactly.
\end{enumerate}
then we must necessarily have $\Psi \equiv \cG$.
\end{theorem}

For a fully rigorous version, we refer to Theorem \ref{thm:sc0} in the next Section \ref{sec:rigorous}, which contains quantitative estimates for the approximation error $\Psi \approx \cG$, by decomposing it into \emph{(1)} an error on the near-constant training distribution, \emph{(2)} a boundary condition error, and \emph{(3)} a self-consistency error. 

We now outline the proof of Theorem \ref{thm:sc00}.
\begin{proof}
    We use an overlapping partition of the domain $\Omega$ into subdomains and zoom in. Suppose $\Omega=\cup_{i\in I}\Omega_i$ is an overlapping partition of $\Omega$, such that each one of $\Omega_i$ is a rescaling and shifting of $\Omega$ and is of size $h$. For sufficiently small $h$, the coefficient $a$ is almost constant in each one of the $\Omega_i$. Consider a partition of unity $1=\sum_{i\in I}\chi_i$ such that $\chi_i$ has support in $\Omega_i$. By the assumed exactness for near-constant inputs and scale-consistency \eqref{eq:scaling}, we know that $\Psi$ is exact when restricted to $\Omega_i$. Thus we have by the weak formulation that $$(a\nabla\Psi,\nabla v_i)=(a\nabla \cG,\nabla v_i)=0$$ for any $v_i$ supported in $\Omega_i$. Therefore for any $v$ supported in $\Omega$, we can take $v_i=\chi_i v$ and summing up the weak formulation for all $i$ and arrive at  
    \[
    (a\nabla\Psi,\nabla v)=\sum_{i\in I} (a\nabla \Psi,\nabla v_i)=\sum_{i\in I} (a\nabla \cG,\nabla v_i)=0.
    \]
    Therefore $\Psi$ is a weak solution with the desired boundary condition, and thus $\Psi=\cG$.
\end{proof}

\subsection{Quantitative estimates when $\Psi$ is only approximately self-consistent.}
\label{sec:rigorous}

In practice, the trained neural operator $\Psi$ cannot be exactly self-consistent, and will also not be exact on near-constant input functions. At best, $\Psi$ can be trained to achieve a small self-consistency error and a small supervised error on a training set of simple input functions. 

It is therefore desirable to have a more quantitative result, providing a rigorous bound on the out-of-distribution error in terms of error on a training set and the self-consistency error of the trained model. Such an extension is achieved in Theorem \ref{thm:sc0}, below. There, we show that the out-of-distribution error on the test distribution can be bounded by a sum of (1) the error on the training distribution, (2) the boundary condition error and (3) the self-consistency error. Before stating this rigorous bound, we introduce the relevant training and test sets, $\mathcal{D}_\delta$ and $\mathcal{D}_M$, as well as defining the relevant errors.

\paragraph{Training data $\mathcal{D}_\delta$ and test data $\mathcal{D}_M$.}
Given the constraint \eqref{eq:coercive}, we now consider a training dataset $\mathcal{D}_\delta$, consisting of nearly constant input data, and a test set $\mathcal{D}_M$ consisting of far-from-constant inputs. 

To this end, we define $\mathcal{D}_s$ for general $s>0$ as follows:
\[
\mathcal{D}_s
=
\left\{
(a,g) 
\, | \,
\Vert \nabla a \Vert_{L^\infty(\Omega)} \le s, 
\text{ and $(a,g)$ satsify \eqref{eq:coercive}}
\right\}.
\]

For small $\delta > 0$, it is clear that any coefficient field $a(x)$ belonging to $\mathcal{D}_\delta$ is nearly constant (having only variations of size at most $\delta$). We will assume that $\Psi$ is trained on such ``simple'' training data $\mathcal{D}_\delta$ for $\delta \ll 1$. We will test $\Psi$ on $\mathcal{D}_M$ for large $M\gg 1$. Clearly, when $M > \delta$, this is an out-of-distribution task, requiring strong generalization. 

\paragraph{Error over $\mathcal{D}_s$.}
To allow quantitative error estimates which measure the generalization capability of a neural operator $\Psi$, we introduce the following error:
\begin{align*}
\mathrm{Err}_{\mathcal{D}_s}(\Psi)
:=
\sup_{(a,g) \in \mathcal{D}_s}
\Vert \Psi(a,g) - \cG(a,g) \Vert_{L^2(\Omega)}.
\end{align*}
Clearly, since $\mathcal{D}_\delta \subset \mathcal{D}_M$, there is a trivial bound
$\mathrm{Err}_{\mathcal{D}_\delta}(\Psi)
\le \mathrm{Err}_{\mathcal{D}_M}(\Psi)$, valid for \emph{any} $\Psi$.
Our goal in the following is to instead derive an estimate in \emph{the non-trivial direction}; we aim to estimate $\mathrm{Err}_{\mathcal{D}_M}(\Psi)$, i.e. the error over the more complicated test distribution $\mathcal{D}_M$, in terms of $\mathrm{Err}_{\mathcal{D}_\delta}(\Psi)$, i.e. the error over the training distribution $\mathcal{D}_\delta$ consisting of nearly constant coefficient fields. In this case, since $\mathcal{D}_M \not \subset \mathcal{D}_\delta$, there is no trivial way to bound $\mathrm{Err}_{\mathcal{D}_M}(\Psi)$ in terms of $\mathrm{Err}_{\mathcal{D}_\delta}(\Psi)$, and we will require self-consistency to fill this gap. 

Therefore, we show rigorously that \emph{self-consistency enables generalization from training on simple inputs to out-of-distribution testing on complex inputs}.

\paragraph{Self-consistency and boundary condition errors.}
In addition to the test and training errors above, we also introduce the boundary condition error
\begin{align}
\label{eq:boundary-error}
\mathrm{Err}_{\mathrm{boundary}}(\Psi)
:=
\sup_{(a,g)\in \mathcal{D}_M}
\Vert 
\Psi(a,g)|_{\partial \Omega} - g
\Vert_{L^2(\partial \Omega)}.
\end{align}
And finally, the following self-consistency error, for $\lambda := \delta / M$,
\begin{align}
\label{eq:selfconsistency-error}
\mathrm{Err}_{\mathrm{selfcon.}}(\Psi)
:=
\sup_{(a,g)\in \mathcal{D}_M} \sup_{\Omega_\lambda \subset \Omega}
\Vert 
\mathcal{T}_\lambda \Psi(a,g) 
- 
\Psi\left(
\mathcal{T}_\lambda a, \mathcal{T}_\lambda \Psi(a,g)|_{\partial \Omega}
\right)
\Vert_{L^2(\Omega)}.
\end{align}
The second supremum in the definition of $\mathrm{Err}_{\mathrm{selfcon.}}$ is over all $\Omega_\lambda\subset \Omega$ of the form $\Omega_\lambda := b + \lambda [0,1]^d$, with corresponding re-scaling  $(\mathcal{T}_\lambda a)(x) = a(\lambda x + b)$. In this supremum, the scaling parameter $\lambda = \delta / M$ is fixed, and we consider all admissible shifts $b\in [0,1-\lambda]^d$, corresponding to the requirement that $\Omega_\lambda \subset \Omega$.

\paragraph{Quantitative estimate for out-of-distribution testing.}

Given the above definitions, we can now provide a more quantitative counterpart to to Theorem \ref{thm:sc} in the main text.

\begin{theorem}
\label{thm:sc0}
Fix $\delta, M>0$, and assume that $M > \delta$. Then there exists a constant $C = C(\delta, M)>0$, such that 
\begin{align}
\label{eq:sc-bound}
\mathrm{Err}_{\mathcal{D}_M}(\Psi)
\le 
C \Big(
\underbrace{
\mathrm{Err}_{\mathcal{D}_\delta}(\Psi)
+
\mathrm{Err}_{\mathrm{boundary}}(\Psi)
}_{
\text{supervised}
}
+
\underbrace{
\mathrm{Err}_{\mathrm{selfcon.}}(\Psi)
}_{
\text{unsupervised}
}
\Big).
\end{align}
In particular, the \emph{out-of-distribution error} $\mathrm{Err}_{\mathcal{D}_M}(\Psi)$ is rigorously bounded by 
\begin{enumerate}
\item the \textbf{error on the training distribution} $\mathrm{Err}_{\mathcal{D}_\delta}(\Psi)$, 
\item the \textbf{boundary condition error} $\mathrm{Err}_{\mathrm{boundary}}(\Psi)$, 
\item and the \textbf{self-consistency error} $\mathrm{Err}_{\mathrm{selfcon.}}(\Psi)$.
\end{enumerate}
\end{theorem}

The simplified version in the main text is obtained when assuming that the supervised and unsupervised contributions in \eqref{eq:sc-bound} vanish, implying that also $\mathrm{Err}_{\mathcal{D}_M} = 0$, i.e. $\Psi(a,g) = \cG(a,g)$ for all $(a,g) \in \mathcal{D}_M$.

Before coming to the proof of Theorem \ref{thm:sc0}, we remark on a (closely related) probabilistic setting, where input functions are drawn from a probability measure $\mu$:
\begin{remark}
For simplicity, our statement of Theorem \ref{thm:sc0} is formulated in terms of sup-errors over the relevant datasets. In principle, the proof could be extended to an alternative setting, where the supremum errors are replaced by average (MSE) errors over suitable training and testing probability measures, whose samples belong to $\mathcal{D}_\delta$ and $\mathcal{D}_M$, respectively. Although this alternative setting is equally relevant, its statement and proof would further complicate the mathematical statement, without providing additional insights. Hence, we have decided to restrict our attention to the sup-error setting.
\end{remark}

We now come to the proof of Theorem \ref{thm:sc0}.
\begin{proof}
We start by fixing $(a,g) \in \mathcal{D}_M$. We note that for $\lambda := \delta /M$ and any $x_0 \in [0,1-\lambda]^d$, the rescaled function $(\T_\lambda a)(\xi) := a(x_0 + \lambda \xi)$ satisfies
\[
\nabla_\xi (\T_\lambda a)(\xi) = \lambda \nabla_x a(x_0 + \lambda \xi),
\]
and hence $\Vert \nabla \T_\lambda a \Vert_{L^\infty(\Omega)}\le \lambda \Vert \nabla a\Vert_{L^\infty(\Omega)} \le \lambda M = \delta$. Thus, $a \in \mathcal{D}_M$ implies $\mathcal{T}_\lambda a \in \mathcal{D}_\delta$ for this choice of $\lambda$. This motivates our definition of $\lambda$, which we fix for the remainder of the proof.

After this simple observation, we now aim to bound $\Vert \Psi(a,g) - \cG(a,g) \Vert_{L^2(\Omega)}$. To this end, first use the triangle inequality to bound,
\[
\Vert \Psi(a,g) - \cG(a,g) \Vert_{L^2(\Omega)}
\le 
\Vert \Psi(a,g) - \cG(a,\Psi(a,g)|_{\partial \Omega}) \Vert_{L^2(\Omega)}
+
\Vert \cG(a,\Psi(a,g)|_{\partial \Omega})- \cG(a,g) \Vert_{L^2(\Omega)}.
\]
The second term arises because $\Psi$ does not necessarily match the boundary conditions perfectly. Since $\cG$ is the true solution operator, it follows from elliptic regularity that the mapping from boundary conditions $g \mapsto \cG(a,g)$ is linear and continuous from $L^2(\partial \Omega) \to L^2(\Omega)$. Hence, there exists a constant $C_0>0$, such that
\[
\Vert \cG(a,\Psi(a,g)|_{\partial \Omega})- \cG(a,g) \Vert_{L^2(\Omega)}
=
\Vert \cG(a,\Psi(a,g)|_{\partial \Omega} - g) \Vert_{L^2(\Omega)}
\le 
C_0 \Vert \Psi(a,g)|_{\partial \Omega} - g \Vert_{L^2(\partial \Omega)}.
\]
In particular, since $(a,g)\in \mathcal{D}_M$, by assumption, the definition of $\mathrm{Err}_{\mathrm{boundary}}$ then implies that 
\begin{align}
\Vert \cG(a,\Psi(a,g)|_{\partial \Omega})- \cG(a,g) \Vert_{L^2(\Omega)}
\le 
C_0 \mathrm{Err}_{\mathrm{boundary}}(\Psi).
\end{align}
To prove the claimed bound \eqref{eq:sc-bound}, it now suffices to show that there exists $C = C(M,\delta)>0$, such that 
\begin{align}
\label{eq:to-show}
\Vert \Psi(a,g) - \cG(a,\Psi(a,g)|_{\partial \Omega}) \Vert_{L^2(\Omega)}
\le 
C \left(
\mathrm{Err}_{\mathcal{D}_\delta}(\Psi)
+
\mathrm{Err}_{\mathrm{selfcon.}}(\Psi)
\right).
\end{align}
This estimate is derived below. To prove it, note that since $\cG$ is the true solution operator, it matches the boundary conditions perfectly. In particular, we have $\cG(a,\Psi(a,g)|_{\partial \Omega})(x) = \Psi(a,g)(x)$ for all $x\in \partial \Omega$. We will use this fact repeatedly in the calculations below, as it implies that several boundary terms vanish when integrating by parts. For the following calculations, we simplify the notation and write in abbreviated form, $\Psi(x) = \Psi(a,g)(x)$ and $\cG(x) = \cG(a,\Psi(a,g)|_{\partial \Omega})(x)$. 

Let now $\phi$ be the unique solution in $H^1_0(\Omega)$ of $-\nabla \cdot (a \nabla \phi) = \Psi - \cG$, $\phi|_{\partial \Omega} = 0$. Then, we have
\[
\int_\Omega |\Psi - \cG|^2 \, dx
=
\int_{\Omega} (\Psi - \cG) (-\nabla \cdot (a\nabla \phi)) \, dx
=
\int_{\Omega} a \nabla (\Psi - \cG) \cdot \nabla \phi \, dx,
\]
where the second equality follows upon integrating by parts and using the fact that $\Psi - \cG$ vanishes on $\partial \Omega$. Since $\phi \in H^1_0$, and since $\cG$ is an exact solution, it follows (again via integration by parts) that 
\begin{align}
\label{eq:phi-trick}
\int_{\Omega}  a\nabla\cG \cdot \nabla \phi \, dx
=
\int_{\Omega} (-\nabla \cdot (a\nabla \cG)) \phi \, dx
= 0,
\quad
\forall \, \phi \in H^1_0(\Omega).
\end{align}
Hence, we have $\int_{\Omega} |\Psi - \cG|^2 \, dx = \int_{\Omega} a \nabla (\Psi - \cG)\cdot \nabla \phi \, dx = \int_{\Omega} a \nabla \Psi \cdot \nabla \phi \, dx$. We now find an upper bound on the last term. To this end, choose a smooth partition of unity $1 = \sum_{j=1}^N \chi_j$, such that each $\chi_j$ is compactly supported in the interior of $\Omega_j \subset \Omega$, and $\Omega_j$ is of the form $\Omega_j = x_j + \lambda [0,1]^d$ for the re-scaling factor $\lambda = \delta /M >0$ fixed at the beginning of this proof. In fact, we can always ensure that $\chi_j$ is of the form $\chi_j(x) = \chi((x-x_j)/\lambda)$ for a fixed, smooth function $\chi \ge 0$. In particular, this then implies that 
\[
\Vert \nabla \chi_j \Vert_{L^\infty(\Omega)} \le C_{\chi} \lambda^{-1}, 
\quad \text{with }C_{\chi} \text{ fixed, independent of $\lambda$.}
\]
By construction, we then have $\phi = \sum_{j} \chi_j \phi =\sum_j \phi_j$, where $\phi_j := \chi_j \phi$ has support in $\Omega_j$, and vanishes on $\partial \Omega_j$. We now decompose
\begin{align*}
\int_{\Omega} a \nabla \Psi \cdot \nabla \phi \, dx
&= 
\sum_{j=1}^N \int_{\Omega_j} a \nabla \Psi(a,g) \cdot \nabla \phi_j \, dx
\end{align*}
Invoking \eqref{eq:phi-trick} with $\Omega$ replaced by $\Omega_j$, and with $\cG(a|_{\Omega_j},\Psi(a,g)|_{\partial\Omega_j})$ in place of $\cG$, we can further write 
\begin{align*}
\int_{\Omega} a \nabla \Psi \cdot \nabla \phi \, dx
&= 
\sum_{j=1}^N \int_{\Omega_j} a \nabla (\Psi(a,g) - \cG(a|_{\Omega_j},\Psi(a,g)|_{\partial\Omega_j}) ) \cdot \nabla \phi_j \, dx.
\end{align*}
Since $\Psi(a,g) - \cG(a|_{\Omega_j},\Psi(a,g)|_{\partial\Omega_j})$ again vanishes on the boundary of $\Omega_j$, we can integrate by parts to find,
\begin{align*}
\int_{\Omega} |\Psi(a,g) - \cG(a,\Psi(a,g)|_{\partial\Omega})|^2 \, dx
&=
\int_{\Omega} a \nabla \Psi \cdot \nabla \phi \, dx
\\
&= 
\sum_{j=1}^N \int_{\Omega_j} \left(\Psi(a,g) - \cG(a|_{\Omega_j},\Psi(a,g)|_{\partial\Omega_j}) \right) \left(-\nabla \cdot ( a  \nabla \phi_j )\right)\, dx.
\end{align*}
We note that the expression on the left measures $\Vert \Psi(a,g) - \cG(a,\Psi(a,g)|_{\partial \Omega}\Vert_{L^2(\Omega)}^2$, with $\cG(a,\Psi(a,g)|_{\partial \Omega}$ the solution of the elliptic PDE over the whole domain $\Omega$, with coefficient field $a$ and boundary condition $\Psi(a,g)|_{\partial \Omega_j}$, whereas the terms on the right involve $\Psi(a,g)|_{\Omega_j} - \cG(a|_{\Omega_j}, \Psi(a,g)|_{\partial\Omega_j})$, comparing the output $\Psi(a,g)$ to the solution $\cG(a|_{\Omega_j}, \Psi(a,g)|_{\partial\Omega_j})$ of the local elliptic PDE on the subdomain $\Omega_j$, with boundary conditions imposed on boundary of the subdomain $\partial \Omega_j$.

Introducing the following change of variables on $\Omega_j$:  $x = \lambda \xi + x_j$, where $\xi \in [0,1]^d$, and recalling that the relevant re-scaling here is $\T_\lambda a(\xi) = a(\lambda \xi + x_j)$, we can then write,
\begin{align*}
\Psi(a,g)(x) - \cG(a|_{\Omega_j},\Psi(a,g)|_{\partial\Omega_j})(x)
&=
\T_\lambda \Psi(a,g)(\xi) - \Psi(\T_\lambda a, \T_\lambda \Psi(a,g)|_{\partial\Omega})(\xi) \\
&\qquad
+ \Psi(\T_\lambda a, \T_\lambda \Psi(a,g)|_{\partial\Omega})(\xi)
- 
\cG(\T_\lambda a,\T_\lambda\Psi(a,g)|_{\partial\Omega})(\xi)
\end{align*}
The first difference measures the self-consistency error. The second difference is the error between $\Psi$ and $\cG$ on nearly-constant coefficient fields (recall that $\Vert \nabla (\T_\lambda a)\Vert_{L^\infty(\Omega)} \le \delta$ by our choice of $\lambda$). Taking also into account that the integration element $dx = \lambda^d d\xi$, it then follows that
\begin{align*}
\Vert \Psi(a,g) - \cG(a|_{\Omega_j},\Psi(a,g)|_{\partial\Omega_j}) \Vert_{L^2(\Omega_j)}
&\le
\lambda^{d/2} 
\Vert 
\T_\lambda \Psi(a,g) - \Psi(\T_\lambda a, \T_\lambda \Psi(a,g)|_{\partial\Omega})
\Vert_{L^2(\Omega)}
\\
&\qquad
+
\lambda^{d/2}
\Vert 
\Psi(\T_\lambda a, \T_\lambda \Psi(a,g)|_{\partial\Omega})
- 
\cG(\T_\lambda a,\T_\lambda\Psi(a,g)|_{\partial\Omega})
\Vert_{L^2(\Omega)}
\\
&\le \lambda^{d/2} \mathrm{Err}_{\mathrm{selfcon.}}(\Psi) 
+ \lambda^{d/2} \mathrm{Err}_{\mathcal{D}_\delta}(\Psi).
\end{align*}
A short calculation, based on expanding $\nabla \cdot (a\nabla \phi_j) = \nabla \cdot (a\nabla (\chi_j\phi))$ furthermore shows that
\begin{align*}
\Vert \nabla \cdot (a\nabla \phi_j) \Vert_{L^2(\Omega_j)}
\lesssim
\Vert \chi_j \Vert_{W^{2,\infty}(\Omega)} \Vert a \Vert_{W^{1,\infty}(\Omega)} \Vert \phi \Vert_{H^1(\Omega_j)}
+ \Vert \chi_j \Vert_{L^\infty(\Omega)} \Vert \nabla \cdot (a \nabla \phi) \Vert_{L^2(\Omega_j)}.
\end{align*}
The implied constant here is universal (in fact $2$ would do). Since $\Vert \chi_j \Vert_{W^{2,\infty}} \lesssim \lambda^{-2}$ and $\Vert a \Vert_{W^{1,\infty}} \lesssim M$, and $\Vert \chi_j \Vert_{L^\infty} \le 1$, we obtain an estimate 
\[
\Vert \nabla \cdot (a\nabla \phi_j) \Vert_{L^2(\Omega_j)}
\le 
C_1 \Vert \phi \Vert_{H^1(\Omega_j)} + \Vert \nabla \cdot (a \nabla \phi) \Vert_{L^2(\Omega_j)},
\]
for a constant $C_1 = C_1(M, \delta)>0$ depending only on $M$ and $\delta$ (through $\lambda = \delta / M$).
Employing these estimates on $\Psi-\cG$ and $\nabla \cdot (a\nabla \phi_j)$, we obtain, using Cauchy-Schwarz inequality and then applying $ab \le \frac{s}{2} a^2 + \frac{1}{2s} b^2$ for arbitrary $s > 0$ to the product of $L^2$-norms, we can bound,
\begin{align*}
&\sum_{j=1}^N \int_{\Omega_j} \left(\Psi(a,g) - \cG(a|_{\Omega_j},\Psi(a,g)|_{\partial\Omega_j}) \right) \left(-\nabla \cdot ( a  \nabla \phi_j )\right)\, dx 
\\
&\qquad
\le
\sum_{j=1}^N s \lambda^d \left\{ \mathrm{Err}_{\mathrm{selfcon.}}(\Psi)^2 
+ \mathrm{Err}_{\mathcal{D}_\delta}(\Psi)^2\right\}
+
\sum_{j=1}^N \frac{1}{s} \left\{
C_1^2 \Vert \phi \Vert_{H^1(\Omega_j)}^2 + \Vert \nabla \cdot (a \nabla \phi) \Vert_{L^2(\Omega_j)}^2
\right\}
\\
&= s (N\lambda^d) \left\{ \mathrm{Err}_{\mathrm{selfcon.}}(\Psi)^2 
+ \mathrm{Err}_{\mathcal{D}_\delta}(\Psi)^2\right\}
+ 
\frac{1}{s} \left\{
C_1^2 \Vert \phi \Vert_{H^1(\Omega)}^2 + \Vert \nabla \cdot (a \nabla \phi) \Vert_{L^2(\Omega)}^2
\right\}.
\end{align*}
We  note that for a suitable partition of unity $\chi_j$, we can always ensure that only adjacent domains $\Omega_j$ overlap (leading to a maximal overlap of $2^d$ domains near the corners, with dimension $d \in \{1,2,3\}$ fixed), implying that $N\lambda^d$ is no greater than $2^d$ times the measure of the covered domain $\Omega = [0,1]^d$. Hence $N\lambda^d\le C_d = 2^d$ is uniformly bounded by a constant depending only on $d$; in fact, $C_d \le 8$ for the most relevant $d\in \{1,2,3\}$.

Since $\phi$ by definition solves $-\nabla (a\nabla \phi) = \Psi(a,g) - \cG(a,\Psi(a,g)|_{\partial \Omega})$, it follows from the theory of elliptic PDEs that 
\[
\Vert \phi \Vert_{H^1(\Omega)}, 
\, \Vert \nabla \cdot (a \nabla \phi) \Vert_{L^2(\Omega)}
\le 
C_2 \Vert \Psi(a,g) - \cG(a,\Psi(a,g)|_{\partial \Omega}) \Vert_{L^2(\Omega)},
\]
for some constant $C_2$ depending on $d$ and the domain $\Omega = [0,1]^d$, which is fixed.

Combining these estimates, and choosing the free parameter $s>0$ to balance terms, we (finally!) conclude that for some constant $C = C(M,\delta,d)>0$, we have
\begin{align*}
\Vert \Psi(a,g) - \cG(a, \Psi(a,g)|_{\partial \Omega}) \Vert_{L^2(\Omega)}^2
&= \sum_{j=1}^N \int_{\Omega_j} \left(\Psi(a,g) - \cG(a|_{\Omega_j},\Psi(a,g)|_{\partial\Omega_j}) \right) \left(-\nabla \cdot ( a  \nabla \phi_j )\right)\, dx 
\\
&\le\sqrt{ C_d \left\{\mathrm{Err}_{\mathrm{selfcon.}}(\Psi)^2 
+ \mathrm{Err}_{\mathcal{D}_\delta}(\Psi)^2\right\}}
\sqrt{C_1^2\Vert \phi \Vert_{H^1(\Omega)}^2 + \Vert \nabla \cdot (a \nabla \phi) \Vert_{L^2(\Omega)}^2}
\\
&\le 
C \left\{\mathrm{Err}_{\mathrm{selfcon.}} (\Psi)
+ \mathrm{Err}_{\mathcal{D}_\delta}(\Psi)\right\}
\Vert \Psi(a,g) - \cG(a,\Psi(a,g)|_{\partial \Omega} \Vert_{L^2(\Omega)}.
\end{align*}
Hence, 
\begin{align*}
\Vert \Psi(a,g) - \cG(a, \Psi(a,g)|_{\partial \Omega}) \Vert_{L^2(\Omega)}
\le 
C \left\{\mathrm{Err}_{\mathrm{selfcon.}}(\Psi) 
+ \mathrm{Err}_{\mathcal{D}_\delta}(\Psi)\right\}.
\end{align*}
This is inequality \eqref{eq:to-show}, which remained to be shown, and concludes our proof.
\end{proof}

%% file: Sections/A-Implementation.tex
\section{Implementation Details}
The overall architecture is shown in Figure \ref{fig:architecture}.

\subsection{Fourier Neural Operator}
The neural operator, proposed in \citep{li2020neural}, is formulated as an iterative architecture $f_0 \mapsto f_1 \mapsto \ldots \mapsto f_T$, where $f_j$ for $j=0,1,\dots,T-1$ is a sequence of functions, each taking values in $\mathbb{R}^{C}$. The input $a \in \mathcal{A}$ is first lifted to a higher-dimensional representation $f_0(x) = P(a(x))$ by the local transformation $P$, which is usually parameterized by a shallow fully-connected neural network. The output $u(x) = Q(f_T(x))$ is the projection of $f_T$ by the local transformation $Q: \mathbb{R}^{C} \to \mathbb{R}^{d_u}$. In each iteration, the update $f_t \mapsto f_{t+1}$ is defined as the composition of a non-local integral operator $\mathcal{K}$ and a local, nonlinear activation function $\sigma$.
\begin{equation}
    \cG_{\theta} \coloneqq \mathcal{Q} \circ(W_{L} + \mathcal{K}_{L}) \circ \cdots \circ \sigma(W_1 + \mathcal{K}_1) \circ \mathcal{P}
\end{equation}

Denote the layer $\sigma(W_l + \mathcal{K}_l)$ mapping the representation $f_t \mapsto f_{t+1}$ by
\begin{equation}\label{def:int}
f_{t+1}(x) := \sigma\Big( W f_t(x) 
+ \bigl(\mathcal{K}(a;\phi)f_t\bigr)(x) \Big),
\end{equation}
where $\mathcal{K}$ maps to bounded linear operators on $\U(D; \mathbb{R}^{C})$ and is parameterized by $\phi \in \Theta_{\mathcal{K}}$, $W: \mathbb{R}^{C} \to \mathbb{R}^{C}$ is a linear transformation, and $\sigma : \mathbb{R} \to \mathbb{R}$ is a non-linear activation function whose action is defined component-wise. 

In FNO, the kernel integral operator in $\mathcal{K}$ is defined as a convolution operator in Fourier space. Let $\mathcal{F}$ denote the Fourier transform of a function $f: D \to \mathbb{R}^{C}$ and $\mathcal{F}^{-1}$ its inverse, then
\begin{align*}
    \hat{f}(k) &= (\mathcal{F} f)_j(k) = \int_{D} f_j(x) e^{- 2i \pi \langle x, k \rangle} \mathrm{d}x,\\
    f(x) &= (\mathcal{F}^{-1} f)_j(x) = \int_{D} \hat{f}_j(k) e^{2i \pi \langle x, k \rangle} \mathrm{d}k,
\end{align*}

\subsection{Weight Sharing Parameterization}

The spectral convolution is defined as
\begin{equation}
\label{eq:Fourier}
\bigl(\mathcal{K}(\phi)f_t\bigr)(x)=   
\mathcal{F}^{-1}\Bigl(R_\phi \cdot (\mathcal{F} f_t) \Bigr)(x) \qquad \forall x \in D,
\end{equation}
where $R_\phi$ is the learnable weight matrix or weight tensor.

\paragraph{Weight Tensor parameterization.}
Assuming the domain $D$ is discretized with $n \in \mathbb{N}$ points, we have $f_t \in \mathbb{R}^{n \times C}$ and $\mathcal{F} (f_t) \in \mathbb{C}^{n \times C}$. Since we convolve $f_t$ with a function that only has $M_{\text{max}}$ Fourier modes, we may simply truncate the higher modes to obtain $\mathcal{F} (f_t) \in \mathbb{C}^{M_{\text{max}} \times C}$, where $M_{\text{max}} = M_1 \times \ldots \times M_d$. Multiplication by the weight tensor $R \in \mathbb{C}^{M_{\text{max}} \times C \times C}$ is defined as
\begin{equation}
\label{eq:tensor}
\bigl( R \cdot (\mathcal{F} f_t) \bigr)_{k} = \sum_{j=1}^{C} R_{k,j}  (\mathcal{F} f_t)_{k,j}.
\end{equation}

\paragraph{Weight sharing parameterization.}
Multiplication by the weight matrix $R \in \mathbb{C}^{C \times C}$ is defined as
\begin{equation}
\label{eq:matrix}
\bigl( R \cdot (\mathcal{F} f_t) \bigr)_{k} = \sum_{j=1}^{C} R_{j}  (\mathcal{F} f_t)_{k,j}.
\end{equation}
For the matrix parameterization, it is optional to add a bias term $b \in \mathbb{C}^{C}$.

\paragraph{Combining matrix and tensor parameterization.}
The multi-band structure is designed in a robust manner, allowing the specification of the channel dimension $C_l$ and bandwidth $M_l$ to any size. It is also flexible to combine the tensor parameterization \eqref{eq:tensor} and matrix parameterization \eqref{eq:matrix}. In practice, we use the first level as weight-sharing parameterization to have a full-frequency convolution, and the rest of the levels as tensor versions.

\subsection{Frequency Encoding}
The wavenumber $k \in \mathbb{Z}$ is encoded to a frequency feature $\mathbb{C}^{C}$ by a frequency encoding layer before being fed into the kernel network. For $C$ channels, we define
\begin{equation}
    k_j = k^{\frac{i}{(C-1)}}, \qquad j = 0,1,\ldots C-1.
\end{equation}
We note that $k_j$ is unbounded and can become very large. As $k \rightarrow \infty$, $k_j \rightarrow \infty$. Since the input signal decays exponentially, $\hat{f}_t(k) = O(\exp (-\alpha k))$, a larger feature will help the model capture smaller signals.

\subsection{Multi-band U-shape architecture}
\label{sec:ushape}
The U-shape architecture consists of down blocks and up blocks.

\textbf{Down Blocks.}
At each level, the input tensor is transformed into shape $(B, C_l, M_{1,l}, \ldots, M_{d,l})$ with the down blocks. The down block consists of two steps: (1) Truncation: Truncate the modes from $M_l$ to $M_{l+1}$. and (2) $\mathcal{K}$ Layer: Apply $R_{l,l+1}$ to lift the channel dimension from $C_l$ to $C_{l+1}$, followed by a complex activation function.
After reaching the lowest level, we have collected the input $\{f, f_1, \ldots, f_L \}$.

\textbf{Up Blocks.}
Conversely, the up blocks lift the tensor back to the original shape. Similarly, it consists of two steps: (1) $\mathcal{K}$ Layer: Apply $R_{l, l-1}$ to project the channel dimension from $C_{l}$ to $C_{l-1}$, followed by a complex activation function. (2) Summation: Combine the output of mode $M_{l}$ with the inputs $f_l$ of $M_{l-1}$ by adding corresponding modes.

\textbf{Skip Connection.}
Furthermore, we define skip connections in the Fourier space. After the down block, we save the intermediate tensors $\{f, f_1, \ldots, f_L \}$ and pass them to the next layer. The skip-out tensor at layer $t$ will be added back at the next layer $t+1$ in the down block.

\subsection{Activation Functions on Complex Space}
\label{sec:complex_activation}
$R$ is a complex kernel neural network $R: \mathbb{C}^{C_{in} + C} \to \mathbb{C}^{C_{out}}$. We use a complex GeLU as the activation function, which applies GeLU to the real and imaginary parts separately, similar to the complex ReLU in \cite{trabelsi2017deep}. This choice empirically provides the best performance.
\begin{equation}
    \text{cGeLU}(\hat{f}): \text{GeLU}(\text{real}(\hat{f})) + i \text{GeLU}(\text{imag}(\hat{f})).
\end{equation}

%% file: Sections/A-Experiments.tex
\begin{table}
    \centering
    \caption{Navier-Stokes equation trained on RE1000, zero-shot test on various RE (2+1 dimensional models).}
    \label{table:NS-spacetime}
    \resizebox{\textwidth}{!}{%
    \begin{tabular}{lcccc|ccc|ccc}
        \toprule
        \textbf{Model} & \textbf{Scale} & \textbf{Freq.} & \textbf{Aug.} & \textbf{size} & \textbf{Re=250} & \textbf{Re=500} & \textbf{Re=1000} & \textbf{Re=2000} & \textbf{Re=4000} & \textbf{Re=10000} \\
        & \textbf{Informed} & \textbf{Emb.} & & \textbf{min} & \textbf{256} & \textbf{256} & \textbf{512} & \textbf{512} & \textbf{1024} & \textbf{1024} \\
        \midrule
        \textbf{2+1 dim FNO} & No & No & OFF & N/A & 0.02040 & 0.02901 & 0.04460 & 0.08573 & 0.12081 & 0.19554 \\
        & No & Yes & OFF & N/A & 0.01727 & 0.02632 & 0.04051 & 0.08158 & 0.11603 & 0.18847 \\
        & Yes & No & OFF & N/A & 0.01937 & 0.02779 & 0.04194 & 0.08319 & 0.11889 & 0.19588 \\
        & Yes & Yes & OFF & N/A & 0.01756 & 0.02551 & 0.04003 & 0.08029 & 0.11274 & 0.18551 \\
        \midrule
        \textbf{2+1 dim FNO MLP} & No & No & OFF & N/A & 0.03945 & 0.0543 & 0.06768 & 0.11215 & 0.14987 & 0.21862 \\
        & No & Yes & OFF & N/A & 0.03586 & 0.04348 & 0.02827 & 0.06307 & 0.15211 & 0.23422 \\
        & Yes & No & OFF & N/A & 0.04032 & 0.05580 & 0.06803 & 0.11358 & 0.16708 & 0.23437 \\
        & Yes & Yes & OFF & N/A & 0.02125 & 0.02661 & \textbf{0.02701} & \textbf{0.06164} & 0.11917 & 0.19405 \\
        \midrule
        \textbf{2+1 dim FNO} & Yes & Yes & ON & 24 & 0.01352 & 0.02082 & 0.03547 & 0.07420 & 0.11074 & 0.18526 \\
        & Yes & Yes & ON & 32 & \textbf{0.01342} & \textbf{0.02016} & 0.03382 & 0.07285 & 0.10876 & 0.18469 \\
        & Yes & Yes & ON & 40 & 0.01468 & 0.02031 & 0.03348 & 0.07083 & 0.10444 & 0.17692 \\
        & Yes & Yes & ON & 48 & 0.01756 & 0.02515 & 0.03869 & 0.07732 & 0.11194 & 0.18408 \\
        \midrule
        \textbf{2+1 dim FNO MLP} & No & No & ON & 32 & 0.04083 & 0.05681 & 0.06516 & 0.10959 & 0.15138 & 0.22287 \\
        & No & Yes & ON & 32 & 0.01419 & 0.02157 & 0.02917 & 0.06323 & \textbf{0.09880} & \textbf{0.17095} \\
        & Yes & No & ON & 32 & 0.06584 & 0.06874 & 0.06802 & 0.13861 & 0.22819 & 0.35919 \\
        & Yes & Yes & ON & 32 & 0.01750 & 0.02457 & 0.02863 & 0.06271 & 0.13394 & 0.23217 \\
        \bottomrule
    \end{tabular}%
    }
\end{table}

\section{Experimental Details}


\subsection{Scale Consistency Loss}
\label{app:exp}

We test scale consistency on the Darcy Flow, Helmholtz equation, and Navier-Stokes equation.

\paragraph{Darcy Flow}
In the Darcy Flow problem, since the solution is smooth and low-frequency, we use FNO as the baseline. As the domain is not periodic, we use domain padding similar to \cite{liu2023spfno} and normalize the model output by the magnitude of boundary inputs, as discussed in Section \ref{sec:model}. We use 20 Fourier modes, a width (channel dimension) of 64, and 4 layers for the runs with or without self-consistency. The super-sampling has an annealed learning rate with respect to the epoch, where we multiply the rate learn by $\alpha = ep/ep_{\max}$, where $ep = 0, 1, \ldots, ep_{\max}$.

\paragraph{Helmholtz Equation}
For the Helmholtz equation, we compare FNO with the scale-informed FNO. Again, we normalize the model output by the magnitude of boundary inputs. Since the Helmholtz equation has higher frequency components, we use 64 Fourier modes, a width (channel dimension) of 32, and 4 layers. We use an annealed learning rate $\alpha = ep/ep_{\max}$ for super-sampling.

\paragraph{Navier-Stokes Equation}
For the Navier-Stokes equation, we compare UNet, FNO, and the scale-informed neural operator. For UNet, we set 5 levels with channels ranging from 64 to 1024. For FNO, we set 32 Fourier modes and a width (channel dimension) of 32. For the scale-informed neural operator, we also set 32 Fourier modes and a channel dimension of 32, where the first level is MLP-based and the second level is tensor-based.

\paragraph{Minimum size in sub-domain sampling}.
While Algorithm \ref{algo:sub-sample} allows arbitrary subdomains, in practice we need to set a minimum resolution for the problem to make sense. If the subdomain is too small, for example, containing only one pixel, then there is no information contained in the sub-domain problem. We conduct an ablation study on the minimum size for sub sampling on the Navier-Stokes equation. As shown in Table \ref{table:NS-spacetime}, a minimum resolution of $32$ per each dimension of space and time works the best.


\subsection{Spatiotemporal Models}
\begin{figure}[t]
\centering
\includegraphics[width=0.6\columnwidth]{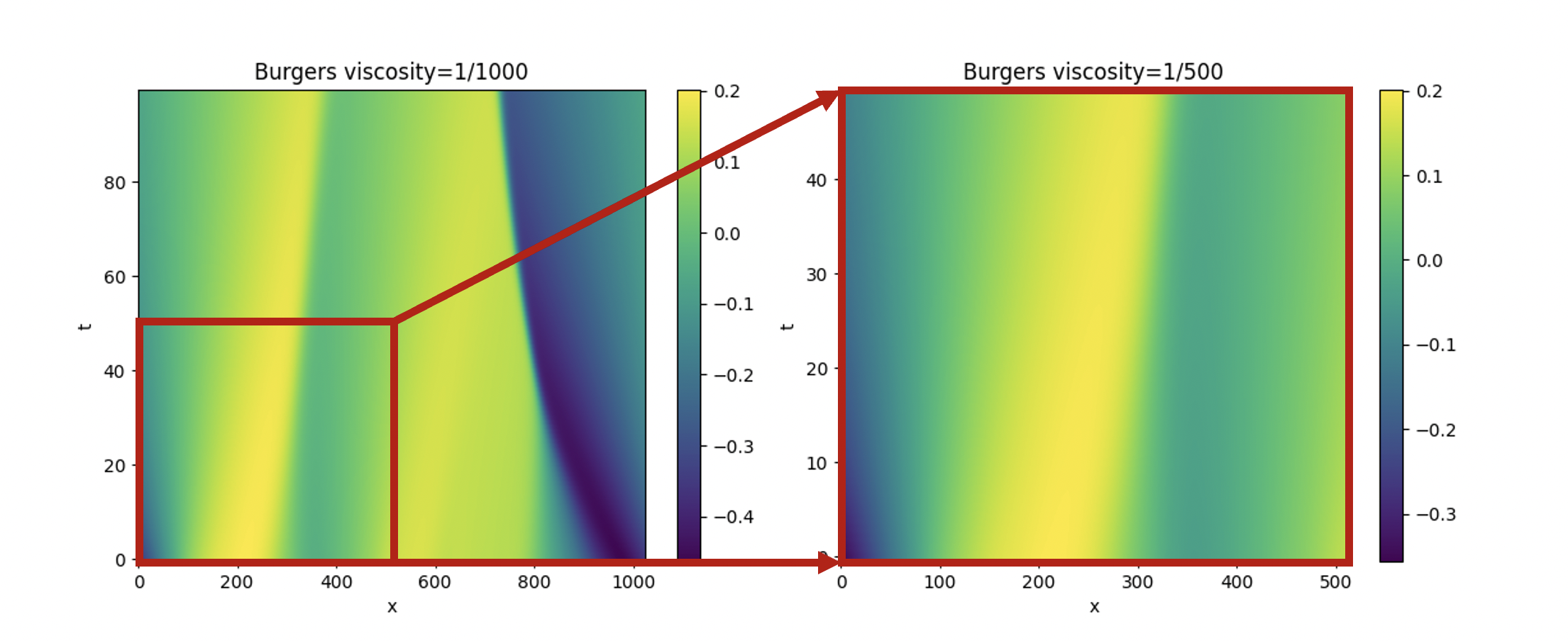}
\caption{Enforcing scale-consistency on Burgers' equation. For time-dependent problem, we treat the time dimension as another spatial dimension. The initial condition and time-dependent boundaries are given to the sub-domain model.}
\label{fig:burgers}
\end{figure}
Many PDEs of practical interest, such as the Navier--Stokes equations, evolve in both space and time. While the previous subsections focus on steady-state or purely spatial formulations, we now extend our framework to a 2+1 dimensional representation. For Navier--Stokes on a 2D spatial domain with time horizon $T$, the data becomes $\{x(\mathbf{x}, t), y(\mathbf{x}, t)\}$ defined over $\Omega \times [0,T]$. We discretize this domain on a grid of size $N_x \times N_y \times N_t$ by lifting the solution into three dimensions. 


\paragraph{Self-consistency via Spatiotemporal Crops.}
For time-dependent PDEs, we split the spatiotemporal domain into two halves along the temporal axis. The model input contains the known solution fields $\mathbf{u}(\cdot, t < t_{\mathrm{mid}})$ with selected boundary information over $\Omega \times [0,T]$. In Navier--Stokes, we pass the velocity field in the first 24 time steps as an input channel and also embed boundary conditions from the last 24 time steps. The operator then predicts the solution in the second half of the time domain. We implement this by concatenating internal state variables and boundary conditions along the channel dimension.

Even with a single training scale (e.g., $Re=1000$), we can enforce scale-awareness and sub-domain consistency in 3D by cropping smaller 3D blocks (subdomains in space and sub-intervals in time). We randomly pick a sub-domain $
\hat{\Omega} \times [t_1,\, t_2]
~\subset~
\Omega \times [0,T], $
with $\hat{\Omega} = [x_0,\, x_0+\hat{h}] \times [y_0,\, y_0+\hat{w}]$, and choose $[t_1,\, t_2]$ so that the sub-domain in time is symmetric around the boundary separating the first $T/2$ timesteps and the latter $T/2$ timesteps. The original Reynolds number $Re$ is then scaled to $\widetilde{Re} = \lambda \cdot Re$ by $
\lambda 
= 
\sqrt[\,3]{\,\tfrac{\hat{h}\,\hat{w}\,(t_2-t_1)}{\,H\,W\,T\,}}.$

Furthermore, restricting the minimum size of the sub-domain sampled to the full length of the temporal window $T=48$ disallows partial sub-intervals in time, preventing cropping from any spatiotemporal reductions. This shows that the zero-shot $Re$ performance on lower $Re$ is clearly amplified by sub-sampling in the 2+1 dimensional models and not because of an increase in training data from additional augmentation.

\subsection{Cost versus Accuracy Study}

We assess the trade-off between computational cost and accuracy by comparing the performance of various models on the Navier-Stokes flow with $Re=5000$ to our baseline models at various memory consumption levels. Our comparison metric is the relative L2 loss, recorded after 50 epochs. We use the maximum number of modes for each model and vary the channel dimensions.

The results, as detailed in Figure \ref{fig:ablation}, demonstrate that the proposed model shows superior performance, particularly at larger widths. Notably, the model can match the performance of FNO with one-tenth the number of parameters and exceeds the performance of the U-shaped variants by more than 15\%, especially at higher memory consumption levels.